\def\arxivbuild{1}
\documentclass{article}
\usepackage[T1]{fontenc}
\usepackage{iclr2026_delta,times}
\ifdefined\arxivbuild
    \iclrpreprintcopy
\fi
\usepackage{amsmath,amssymb,graphicx,booktabs}
\usepackage{arydshln}
\usepackage{hyperref}
\usepackage{url}
\providecommand{\figref}[1]{Figure~\ref{#1}}

\title{On the Generation Order in GFlowNets}
\ifdefined\arxivbuild
\author{%
    Timofei Gritsaev\thanks{Equal contribution.}\\
    Constructor University
    \And
    Evgeny Katasonov\textsuperscript{*}\\
    HSE University
    \And
    Ildus Sadrtdinov\\
    Constructor University
    \AND
    \hspace*{0.8in}Esmeralda S. Whitammer\\
    \hspace*{0.8in}University of Edinburgh
    \And
    Dmitry Vetrov\\
    Constructor University
}

\else
    \author{Anonymous authors}
\fi
\begin{document}
\maketitle
\begin{abstract}
Sampling from discrete distributions specified by unnormalised probability functions is a classical statistical problem. GFlowNets address this problem by constructing objects sequentially, adding one discrete component at a time in either a predefined or a learned order. The choice of order determines which conditional distributions the neural network must learn, some of which may be easier to learn than others. Yet whether existing GFlowNet algorithms discover interpretable generation orders and make use of them during training remains unexplored. We introduce controlled problems where generation order is experimentally shown to affect convergence and standard GFlowNet training struggles to discover useful orders. To address this difficulty, we develop Learning-Order GFlowNet, a training recipe that learns interpretable generation orders without order annotations and improves convergence on these problems. Finally, we investigate gradient signal-to-noise ratio (SNR) as a possible explanation for order emergence: orders associated with faster convergence exhibit higher SNR at comparable validation loss, suggesting both a mechanism favouring their emergence and an SNR-based characterisation of a good generation order.
\end{abstract}

\section{Introduction}


We consider the problem of sampling from a discrete space $\mathcal{X}$. Given an unnormalised probability function $R$, also called a reward function, our goal is to sample from
\begin{equation}
    p(x) = \frac{R(x)}{Z}, \qquad Z = \sum_{x \in \mathcal{X}} R(x),
\end{equation}
where $p$ is the target distribution and $Z$ is its normalisation constant. The space $\mathcal{X}$ consists of complete discrete objects, and $Z$ is generally intractable to compute. This is a classical problem with many applications, for example, in drug discovery~\citep{bengio2021flow, cretu2025synflownet} and graph combinatorial optimisation~\citep{zhang2023let}.

Many discrete objects admit a hierarchical decomposition, and GFlowNets~\citep{bengio2021flow, malkin2022trajectory} provide an amortised approach to sampling based on this decomposition. 
Backward, or decomposing, actions define a directed acyclic graph (DAG) in which complete objects from $\mathcal{X}$ are progressively decomposed to an empty state. Reversing these edges gives the forward, or generative, actions 
and associated DAG: objects are constructed from the empty state through a sequence of intermediate states. The backward policy determines the distribution over decomposition orders. It can be fixed, for example by selecting uniformly among removable components or by following a predefined order, as in autoregressive generation. Alternatively, \citet{malkin2022trajectory} parameterise the backward policy alongside the forward policy and learn them jointly.
However, which generation orders emerge from this joint learning, and why they emerge, remain insufficiently understood.

In the generative modelling, it was shown that generation order can change the performance of a trained model~\citep{Vinyals2015OrderMS}. While some domains have common orders, for example, text is often generated from left to right, some domains, like graphs, do not have a preferred ordering~\citep{chen2021order,wang2025learning}. Any-order autoregressive models avoid a fixed order by training over uniformly sampled permutations of the data dimensions~\citep{uria2014deep}, but such order-agnostic models can underperform fixed-order autoregressive models~\citep{hoogeboom2022autoregressive}. A possible explanation is that a neural network is trained to fit conditional distributions across all reveal orders, including difficult prediction problems, without learning a preference for more suitable orders~\citep{wang2025learning,kim2025train}. This motivates learning generation orders that expose tractable conditionals for the target distribution.





In this work, we study interpretable and learnable order in sequential sampling via GFlowNets framework. We outline our main contributions as follows:
\begin{itemize}
    \item We introduce discrete sampling problems in which the generation order changes the difficulty of conditional prediction, and study the effect of the choice of generation order on sampler performance.
    \item We introduce and ablate Learning-Order GFlowNet (LO-GFlowNet) -- a training recipe that allows learning generation orders reflecting the underlying dependencies of the problems.
    \item We suggest a possible explanation for the emergence of interpretable orders through the analysis of gradient signal-to-noise ratio (SNR).
\end{itemize}

\section{Methodology}
In this section, we first introduce a set of novel string problems where varying generation order results in different problem complexity. Next, we remind how a GFlowNet is usually trained and introduce novel tricks leading to order emergence.

\subsection{Problems}

We consider strings $x=(x_1,\ldots,x_L)$ over an alphabet $\mathcal{A}=\{0,\ldots,A-1\}$. A forward action fills one empty position with a symbol, and a backward action erases one filled position.

\paragraph{Markov-chain strings.}
We consider strings drawn from a first-order Markov chain with a high probability of repeating the preceding symbol. The first symbol is uniform, and each subsequent symbol copies its predecessor with high probability $1-\epsilon$; otherwise, it is drawn uniformly from the other $A-1$ symbols:
\begin{equation}
 R(x)=p(x)=\frac{1}{A}\prod_{i=2}^{L}
 \left[(1-\epsilon)\mathbf{1}\{x_i=x_{i-1}\}
 +\frac{\epsilon}{A-1}\mathbf{1}\{x_i\ne x_{i-1}\}\right].
\end{equation}
Generating in left-to-right position order exposes a one-step conditional at every move. Other orders can require predictions conditioned on observations on both sides of an empty position.


\paragraph{Checksum dependencies.}
A string consists of a prefix $u\in\mathcal{A}^{m}$ where each symbol is uniformly sampled from the vocabulary and a dependent suffix $v\in\mathcal{A}^{r}$. For fixed functions $f_j$, the target is
\begin{equation}
    R(x)=p(x=\texttt{concatenate}[u, v])=\frac{1}{A^m}\prod_{j=1}^{r}
    \left[(1-\epsilon)\mathbf{1}\{v_j=f_j(u)\}
    +\frac{\epsilon}{A}\right].
\end{equation}
Similarly to Markov-chain strings, we replace suffix symbols with random ones with probability~$\epsilon$.
In the linear-checksum construction, $f(u)=Wu\bmod A$ for a fixed integer matrix $W$. In the half-sum construction, each $f_j$ sums a specified collection of prefix symbols modulo $A$; for pairwise dependencies, $f_j(u)=(u_{a_j}+u_{b_j})\bmod A$. Generating the required inputs before a suffix symbol turns its prediction into evaluating the corresponding arithmetic rule. 


\paragraph{DAG checksum dependencies.}
We consider a directed acyclic graph $G=(V,E)$, with alphabet
$\mathcal{A}$ and root set $V_0$.
The roots are sampled independently and uniformly. Each other node is assigned
the sum of its parents modulo $A$ with probability $1-\epsilon$;
otherwise, one of the remaining $A-1$ values is chosen uniformly.
Writing
$f_v(x)=\bigl(\sum_{u\in\operatorname{pa}(v)}x_u\bigr)\bmod A$,
we define the reward as
\begin{equation}
\label{eq:dag-reward}
R(x)= p(G) = 
 A^{-|V_0|}
 \prod_{v\in V\setminus V_0}
 \left(
 (1-\epsilon)\,\mathbf{1}\{x_v=f_v(x)\}
 +\frac{\epsilon}{A-1}\,\mathbf{1}\{x_v\neq f_v(x)\}
 \right)
.
\end{equation}
This environment generalizes checksum dependencies from
sequences to directed acyclic graphs. The graph structure is
provided to the sampler, but no generation order is prescribed.

\paragraph{Sudoku.}
We also consider conditional generation of $9\times9$ Sudoku boards. Each puzzle $c$ specifies a set of fixed clues, and the sampler fills the remaining cells with digits from $\{1,\ldots,9\}$. A valid completion contains each digit exactly once in every row, column, and $3\times3$ block. Forward actions choose an empty cell and a digit; backward actions erase a generated digit while leaving the original clues unchanged. Each puzzle $c$ defines a separate conditional target problem over complete boards $x$ consistent with its clues:
\begin{equation}
 p_\beta(x\mid c)=\frac{R_\beta(x,c)}{Z_\beta(c)},
 \qquad R_\beta(x,c)=\exp\!\left[-\beta C(x)\right].
\end{equation}
Here $C(x)$ counts duplicate digits across the 27 rows, columns, and blocks: a digit appearing $k$ times in a unit contributes $\max(k-1,0)$. Consequently, $C(x)=0$ precisely for valid complete boards. Incorrect completions retain a positive reward, providing a graded signal instead of rewarding only fully solved puzzles.

Appendix~\ref{app:implementation_details} includes additional discussion and illustrations of these problems, and environment hyperparameters.

\subsection{Learning-Order GFlowNets}

First, we remind how a GFlowNet can be trained with the loss from ~\citet{malkin2022trajectory} and then show how to modify the recipe so that the model automatically learns an interpretable order related to a target problem. Note that LO-GFlowNet does not use hand-designed labels of any orders, which suggests that it is able to uncover meaningful structures in new problems.

\paragraph{Training a GFlowNet with Trajectory Balance} \citet{malkin2022trajectory} introduce the Trajectory Balance (TB) loss to match forward and backward trajectory distributions:
\begin{equation}
    \mathcal{L}_\text{TB}(\tau = (s_0, s_1, \ldots, s_T)) = \left(\log \left[ \prod_{i=1}^{T} P_F(s_i \mid s_{i-1}, \theta) \right] - \log \left[\frac{R(s_T)}{Z_\theta} \prod_{i=1}^T P_B(s_{i-1} \mid s_i)\right] \right)^2,
    \label{eq:tb}
\end{equation}
where $\prod_{i=1}^{T} P_F(s_i \mid s_{i-1}, \theta)$ is the forward trajectory probability, and $\frac{R(s_T)}{Z_\theta} \prod_{i=1}^T P_B(s_{i-1} \mid s_i)$ is the corresponding backward trajectory weight, with $Z_\theta$ a learned normalising constant. If the expected loss is zero under any full-support trajectory distribution, the forward and backward distributions match, and $Z_\theta$ equals the reward partition function. By the data-processing inequality~\citep{kullback1951information}, the terminal marginals also match, hence, the model samples exactly from the target distribution. For any valid backward policy, there is a unique matching forward policy on states with positive flow~\citep{malkin2022trajectory,bengio2023gflownet}. The backward policy is usually fixed to be uniform or to follow a predefined order; less often, it is also parametrised by $\theta$ and learned jointly with the forward policy. Since GFlowNets are usually trained without a dataset of target samples, $\mathcal{L}_\text{TB}$ is minimised over trajectories sampled from the current forward policy, possibly augmented with epsilon-greedy exploration.

\paragraph{Training choices that promote order emergence.} We find that standard GFlowNet training does not reliably produce structured generation orders. To encourage their emergence, we investigate the following design choices: factorising the forward policy into position and symbol distributions, regularising backward-policy entropy, uniformly initialising the forward-position and backward-policy heads, reducing their learning rates, branching multiple continuations from shared states.

\textbf{Forward policy factorisation.} A common parametrisation of the forward policy predicts an $L \times A$ matrix of logits, masks invalid actions, and applies a softmax over all position--symbol pairs. Following~\citet{emelianenko2019sequence,wang2025learning}, we instead factorise the policy as $p_\theta(\text{symbol},\text{pos}\mid s_t)=p_\theta(\text{pos}\mid s_t)\,p_\theta(\text{symbol}\mid\text{pos},s_t)$. We find that this parametrisation promotes order emergence. In practice, two linear heads predict a length-$L$ vector of position logits and an $L \times A$ matrix of symbol logits. These are normalised separately: over valid positions for the position policy and over symbols at each position for the conditional symbol policy.

\textbf{Backward policy entropy regularisation.} For each valid backward policy, there is a unique matching forward policy on states with positive flow that achieves zero TB loss and samples from the target distribution. Thus, the TB objective alone does not select a particular decomposition order. We seek structured solutions with a consistent choice of which position to erase at each state. To encourage such solutions, we add a backward-policy entropy penalty to the training loss, weighted by $\lambda_{H_\text{B}}$. Minimising this term favours low-entropy, increasingly deterministic backward policies.

\textbf{Uniform order-policy initialisation and reduced learning rates.} We initialise the weights and biases of the final forward-position and backward-policy layers to zero, yielding uniform distributions over valid positions while retaining random initialisation for the symbol head. We also multiply both order-head learning rates by $\eta_{\mathrm{ord}}=0.1$. These choices are intended to prevent arbitrary initial order preferences from being reinforced before the symbol predictor learns useful dependencies. Related techniques have been used previously: \citet{gritsaev2025optimizing} initialise the backward policy uniformly and use a smaller backward-policy learning rate to stabilize GFlowNet training. We extend these choices to the forward-position head of our factorised policy. Uniform-order pretraining has also been used in order-learning models~\citep{Vinyals2015OrderMS,emelianenko2019sequence}; our approach instead starts from uniform position policies and learns them jointly with symbol prediction from the first update, without a separate pretraining phase.


\textbf{Branching continuations.} We sample multiple continuations from shared intermediate states, allowing the model to explore different subsequent position and symbol choices from the same context. Our branching variant uses four descendants per shared prefix while keeping the total number of fresh trajectories per iteration unchanged. Branching is less critical than the other components: our recipe already learns structured orders and often converges without it. 

\section{Experiments}
\label{sec:exp}

\paragraph{Ablation study experiments.} 

Table~\ref{tab:checksum_entropy_ablation} shows that the components of our training recipe contribute jointly to reliable convergence. On checksum dependencies, the full recipe converges in four of five runs, with an average of $72\,900$ iterations when non-converged runs are assigned $100\,000$ iterations. Removing backward entropy regularisation reduces convergence to one of five runs; removing policy factorisation or the reduced order-head learning rate yields two of five converged runs. Uniform initialisation also helps: with the same reduced learning rates, replacing it with the default initialisation reduces convergence from four to three runs and increases the average to $88\,650$ iterations. These comparisons suggest that both the initial order distribution and its rate of adaptation matter.

The optimal entropy coefficient depends on the problem. On checksum dependencies, increasing $\lambda_{H_\text{B}}$ from 1 to 3 reduces the number of converged runs from four to two. On Markov-chain strings, both coefficients yield four converged runs, while $\lambda_{H_\text{B}}=3$ gives a lower average iteration count (Appendix~\ref{app:ablation}, Table~\ref{tab:markov_entropy_ablation}). Branching provides an additional, but less essential, improvement. 

\begin{table}[ht]
    \centering
    \caption{\textbf{Ablation study on Checksum dependencies}. Each configuration uses 5 random seeds. Average iterations include all seeds, assigning 100\,000 iterations to each non-converged run.}
    \label{tab:checksum_entropy_ablation}
    \begin{tabular}{ll@{\hspace{1.5em}}rr}
        \toprule
        \multicolumn{2}{c}{Setup} & Converged & Avg. iterations \\
        \midrule
        w/o logits factorisation & $\lambda_{H_\text{B}}=1,\ \eta_{\mathrm{ord}}=0.1$ & 2/5 & 96\,700 \\
        \midrule
        w/o backward entropy & $\lambda_{H_\text{B}}=0,\ \eta_{\mathrm{ord}}=0.1$ & 1/5 & 95\,000 \\
        \midrule
        w/o reduced order-head lr & $\lambda_{H_\text{B}}=1,\ \eta_{\mathrm{ord}}=1$ & 2/5 & 84\,200 \\
        \midrule
        w/o uniform initialization & $\lambda_{H_\text{B}}=1,\ \eta_{\mathrm{ord}}=1$ & 1/5 & 95\,100 \\
         & $\lambda_{H_\text{B}}=1,\ \eta_{\mathrm{ord}}=0.1$ & 3/5 & 88\,650 \\
        \midrule
        \textbf{Our training recipe} & $\lambda_{H_\text{B}}=1,\ \eta_{\mathrm{ord}}=0.1$ & \textbf{4/5} & 72\,900 \\
         & $\lambda_{H_\text{B}}=3,\ \eta_{\mathrm{ord}}=0.1$ & 2/5 & 77\,050 \\
        \hdashline
        with branching & $\lambda_{H_\text{B}}=1,\ \eta_{\mathrm{ord}}=0.1$ & \textbf{4/5} & \textbf{67\,800} \\
        \bottomrule
    \end{tabular}
\end{table}

Our training recipe learns interpretable generation orders that reflect the dependencies of each problem: locally consecutive generation on Markov-chain strings and a preference for generating prefix symbols early on checksum dependencies. Appendix~\ref{app:ablation} illustrates and quantifies these patterns across ablation configurations and traces their emergence during training. Importantly, the training is split in two phases: an initial order-search phase, followed by learning symbol prediction under a largely stable order. Appendix~\ref{app:ablation} also extends the ablation study to Markov-chain strings, where we observe essentially the same results for each design choice.

\begin{figure}[ht]
\centering
    \caption{\textbf{Results on checksum dependencies.} \textit{Left}: number of converged runs and average number of iterations, assigning 100\,000 iterations to non-converged runs. Fixed orders with $K$ disjoint prefix--suffix swaps, random fixed orders, and uniform $P_B$ use ten seeds each. LO-GFlowNet uses the best configuration from Table~\ref{tab:checksum_entropy_ablation}: our training recipe with branching, $\lambda_{H_\text{B}}=1$, and $\eta_{\mathrm{ord}}=0.1$, evaluated over five seeds. \textit{Right}: ELBO (top) and EUBO (bottom) for $K\in\{0,3,5,7\}$, LO-GFlowNet, and uniform $P_B$. Curves show the 50th percentile and shaded bands the 20th--80th percentiles, computed in log-space for $-\mathrm{ELBO}$ and $\mathrm{EUBO}$, with the ELBO sign restored. 
    }
    \label{fig:checksum_convergence}
    \begin{minipage}[c]{0.39\linewidth}
    \centering
    \small
    \setlength{\tabcolsep}{3pt}
    \begin{tabular}{lcc}
    \toprule
    Sampler & Converged & \shortstack{Avg.\\iterations} \\
    \midrule
    \shortstack[l]{Fixed order\\$K=0$} & \textbf{10/10} & $\mathbf{8\,575}$ \\
    $K=1$ & \textbf{10/10} & $13\,475$ \\
    $K=2$ & 9/10 & $29\,550$ \\
    $K=3$ & 7/10 & $43\,750$ \\
    $K=4$ & 6/10 & $58\,750$ \\
    $K=5$ & 3/10 & $82\,225$ \\
    $K=6$ & 2/10 & $91\,525$ \\
    $K=7$ & 1/10 & $97\,275$ \\
    \hdashline
    \shortstack[l]{Random\\fixed order} & 2/10 & $87\,225$ \\
    \hdashline
    Uniform $P_B$ & 0/10 & $100\,000$ \\
    \midrule
    LO-GFlowNet & 4/5 & $67\,800$ \\
    \bottomrule
    \end{tabular}
    \end{minipage}\hfill
    \begin{minipage}[c]{0.60\linewidth}
    \centering
    \includegraphics[width=\linewidth]{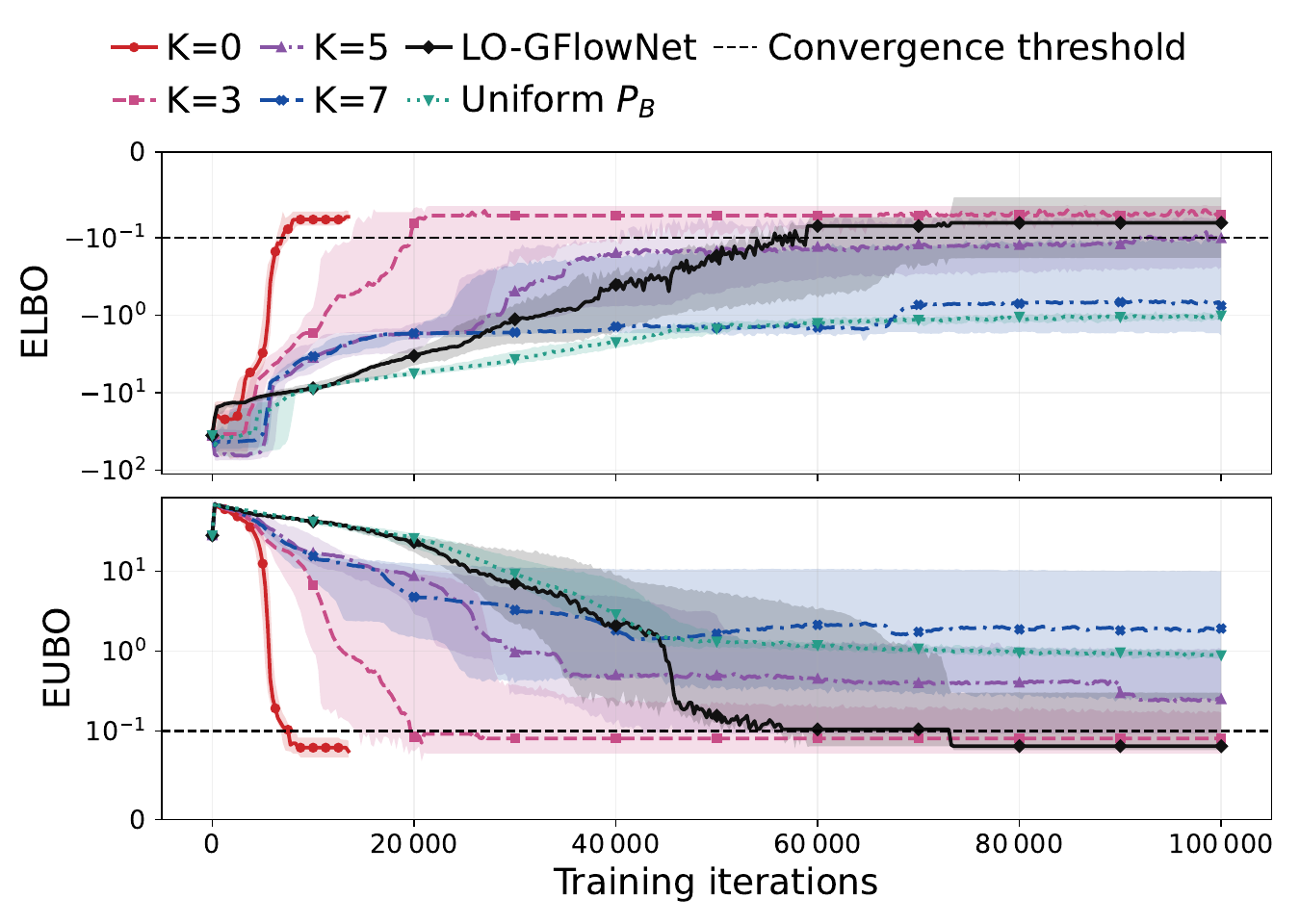}
    \end{minipage}
\end{figure}

\paragraph{Generation order significantly affects convergence.}
For checksum dependencies, we consider GFlowNets with fixed generation orders: $\pi^{\text{CD}}_0$ is the identity permutation, and $\pi^{\text{CD}}_K$ is obtained by applying $K$ swaps, each exchanging a previously unswapped prefix position with a previously unswapped suffix position. As $K$ increases, the fraction of converged runs decreases and the average number of training iterations increases (Figure~\ref{fig:checksum_convergence}). Uniformly sampled fixed random orders converge in only two out of ten runs.

\paragraph{Backward policy choice matters beyond fixed orders.}
A uniform backward policy yields no converged runs within the allocated training budget. This is consistent with the hypothesis that learning symbol distributions across many generation orders is harder than learning those associated with a single fixed order. In contrast, LO-GFlowNet converges in four out of five runs, averaging $67\,800$ iterations when non-converged runs are counted as $100\,000$ iterations. Unlike fixed-order samplers, LO-GFlowNet spends part of this budget learning a generation order, before improving symbol prediction under a largely stable order. Figure~\ref{fig:checksum_order_evolution} shows that it learns structured, prefix-early generation orders, consistent with the dependencies of the problem and the advantage of prefix-first generation observed in the fixed-order experiments.

\paragraph{Our findings extend to graph-generation.}
We next consider DAG checksum dependencies, where the dependencies
between nodes vary across graphs. We compare LO-GFlowNet with an
autoregressive GFlowNet that uses a fixed random generation order.
Both methods use the same network architecture and are trained
on 4,000 graphs, with evaluation on 500 unseen graphs.
LO-GFlowNet outperforms the fixed-order baseline across all five
runs (Table~\ref{tab:dag-results}). This is consistent with the
advantage of generating parents before their children: while a
fixed random order often requires predicting node values before
their parents are known, LO-GFlowNet can adapt its generation
order to the structure of each graph.

\begin{table}[ht]
    \centering
    \caption{\textbf{Results on DAG checksum dependencies.}
    Mean and standard deviation over five seeds, evaluated on
    500 unseen graphs.}
    \label{tab:dag-results}
    \begin{tabular}{lccc}
        \toprule
        Sampler & Converged & ELBO $\uparrow$ & EUBO $\downarrow$ \\
        \midrule
        Random fixed order & 0/5 & $-2.50 \pm 0.02$ & $2.44 \pm 0.05$ \\
        LO-GFlowNet & \textbf{5/5} & $\mathbf{-0.10 \pm 0.03}$ & $\mathbf{0.11 \pm 0.03}$ \\
        \bottomrule
    \end{tabular}
\end{table}


\section{A possible explanation of order emergence: gradients SNR}
\label{sec:snr}

Since a GFlowNet can ocasionally converge even under suboptimal orders, it is not obvious why the training leads to non-random orders.
In this section, we suggest a possible explanation for the emergence of interpretable orders.
Our hypothesis is that some orders expose dependencies that the symbol network can learn more easily through more consistent gradient directions. To measure this, we evaluate gradients SNR (also called GNR in~\citet{fan2022maxgnr}).

We fix a dataset of $N=131\,072$ terminal strings sampled from the exact target distribution and use it to evaluate the validation TB loss.
For fixed orders, each terminal defines a deterministic trajectory; for LO-GFlowNet, we construct a trajectory using its current backward policy.
This validation loss serves as a diagnostic of convergence.
Separately, at each checkpoint saved every 500 iterations, we sample $N$ fresh trajectories from the current forward policy without exploration to measure gradient signal and noise.
We define the reference gradient over the dataset, signal, noise, and SNR, respectively:
\begin{equation*}
    g_{\mathrm{ref}}
    =\frac{1}{N}\sum_{j=1}^{N}\nabla_\theta \mathcal{L}_{\mathrm{TB}}(\tau_j),
    \quad S=\|g_{\mathrm{ref}}\|^2,\quad
    V=\frac{1}{M}\sum_{b=1}^{M}\|g_b-g_{\mathrm{ref}}\|^2,
    \quad \mathrm{SNR}= \sqrt{S/V}.
\end{equation*}
A higher $\mathrm{SNR}$ at a fixed loss indicates for a more consistent gradient direction. Here we independently sample $M=128$ minibatches of 256 trajectories each, with replacement, and $g_b$ is the gradient over the minibatch $b$. Target terminal samples for validation and minibatch index draws are reused across models within each problem; the forward-policy trajectories used for gradient diagnostics are sampled anew at each checkpoint. 

\begin{figure}[ht]
    \centering
    \includegraphics[width=\linewidth]{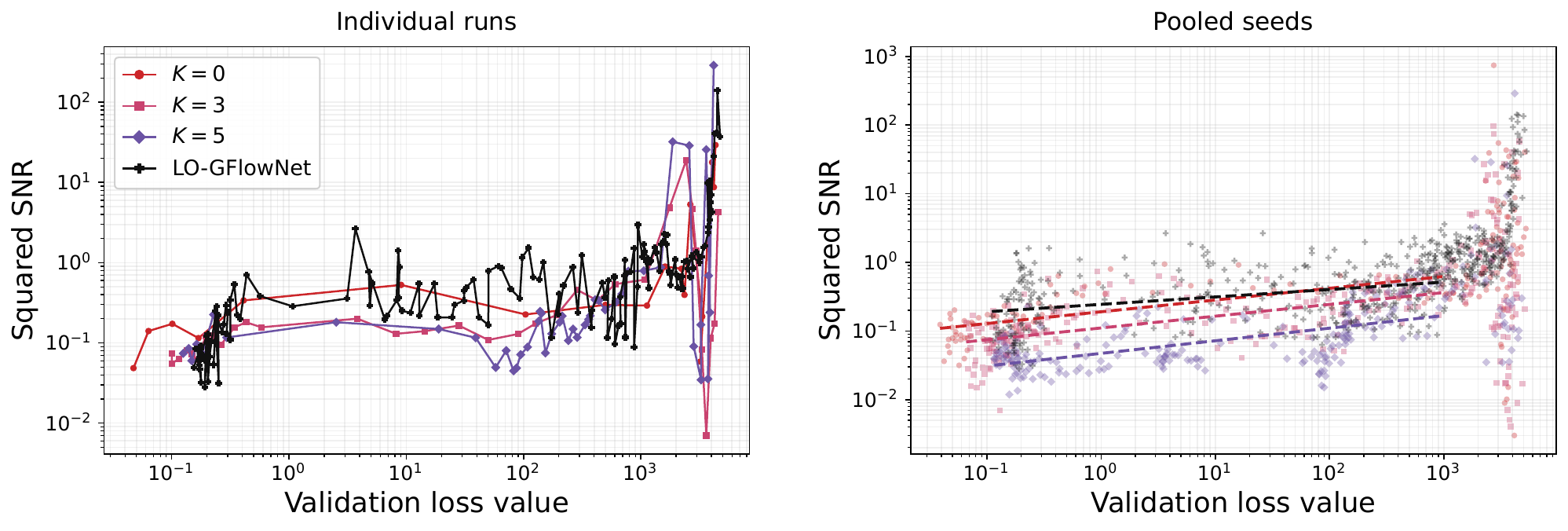}
    \caption{\textbf{Squared gradient SNR on current-policy trajectories versus target-based validation TB loss for checksum dependencies.} Panels show individual runs (left) and pooled measurements (right) for $K=0,3,5$ and LO-GFlowNet, retaining only converged runs. All valid checkpoints are displayed, including those with loss above $10^3$. The left panel uses the first converged seed per configuration. The right panel pools 10, 7, 3, and 4 converged seeds, respectively. Individual checkpoints are connected in training order. Dashed lines are separate least-squares regressions in log--log space, fitted using validation loss below $10^3$.}
    \label{fig:markov_snr_main}
\end{figure}

Figure~\ref{fig:markov_snr_main} compares individual training trajectories and pooled measurements for checksum dependencies.
Among converged runs, LO-GFlowNet reaches SNR comparable to $K=0$, and both show higher SNR than $K=3$ and $K=5$. As demonstrated experimentally in Section~\ref{sec:exp}, increasing $K$ generally produces less useful generation orders in terms of convergence. Together, these observations suggest a more formal characterisation of a good generation order: one that provides a higher gradient SNR at comparable loss values, enabling more consistent updates.

We hypothesise that higher SNR favours the emergence of good orders during training: gradients associated with good orders provide a consistent signal, whereas those associated with poor orders exhibit stronger cancellation, making the useful update harder to distinguish from noise. Individual results for all ten seeds are shown in Appendix~\ref{app:snr}.

\section{Related work}

\paragraph{Generative Flow Networks.} 
\citet{bengio2021flow} introduced GFlowNets trained with a flow matching loss. This formulation does not explicitly parameterise a backward policy and generally admits infinitely many solutions. Later objectives~\citep{malkin2022trajectory,bengio2023gflownet,madan2023learning,tiapkin2024generative,zykova2026proximal} introduce an explicit backward policy that, together with the reward function, defines a target distribution over trajectories for the forward policy to match. For each fixed backward policy, there is a unique matching forward policy on states with positive flow, and these training objectives using a backward policy generally outperform the flow matching loss. Another finding of~\citet{tiapkin2024generative,3702676.3702723} is that GFlowNet training is equivalent to entropy-regularised reinforcement learning (RL) when the backward policy is fixed. From this perspective, the GFlowNet backward policy is equivalent to the RL reward, and the forward policy is an RL agent. A learnable backward policy is a unique feature of GFlowNets that distinguishes it from RL and defines the generation order. It was originally introduced in~\citet{malkin2022trajectory}. Subsequent works~\citep{mohammadpour2024maximum,gritsaev2025optimizing} proposed backward policy training approaches that improve convergence: \citet{mohammadpour2024maximum} derive a backward policy whose matching forward policy maximises trajectory entropy, and \citet{gritsaev2025optimizing} propose a backward policy loss that minimises the KL divergence between forward and backward trajectory distributions, together with practical stabilization techniques that accelerate GFlowNet training. How learned generation orders reflect the target's dependencies, and which mechanisms drive their emergence, remain insufficiently understood. To the best of our knowledge, this work is the first to construct the dicrete sampling problems where order matters, adapt GFlowNets algorithms so that meaningful order emerges and explore why it emerges.

\paragraph{Order in Data-based Generative Models.} Early work studied generation order in \textbf{autoregressive models}. \citet{Vinyals2015OrderMS} show that input and output order can noticeably affect learning in graphical models, combinatorial problems, and language modelling. They also propose likelihood-based output-order search, alternating model-guided selection of output permutations with updates to the sequence model. Crucially, they find that immediately selecting the highest-likelihood order can reinforce an arbitrary initial preference. To mitigate this, they pretrain with uniformly sampled orders and subsequently sample orders according to their likelihood under the current model. These findings highlight the importance of exploration before order specialisation, which is also relevant to our use of uniform position-policy initialisation and stochastic order sampling. Although their fixed-order comparisons cover several domains, their experiments with learned generation orders are limited to five-gram modelling.
\citet{emelianenko2019sequence} introduce an insertion-based sequence model that learns input-dependent generation orders without order annotations, with applications to machine translation, image-to-LaTeX conversion, and image captioning. Their training objective uses a sampled likelihood bound designed to concentrate probability on one or a few convenient insertion trajectories. They find that sampled trajectory selection is more robust than approximate argmax selection, and that pretraining with uniformly sampled valid trajectories improves performance and training robustness. Their analysis is consistent with an easy-first strategy: tokens that are predictable from the input tend to be generated early, while choices requiring more output context are deferred. The preferred orders vary across examples and tasks. 
\citet{chen2021order} treat node order as a latent variable and jointly train an autoregressive graph generator and a variational order distribution by maximising a likelihood lower bound. Their learned order preferences reflect graph structure, including generating communities one at a time.
\citet{wang2025learning} learn flexible generation orders over data dimensions, including both node and edge attributes in molecular graphs. Their model jointly learns a state-dependent position policy and a value predictor by maximising a likelihood lower bound, using a learned auxiliary variational order distribution conditioned on complete training examples. The learned policies exhibit interpretable preferences: generating bonds before atom types in molecular graphs and border pixels before the central digit in MNIST images. Their analysis suggests a preference for high-certainty predictions. They also show that prescribing an edges-first policy with uniform sampling within edge and node groups does not recover the performance of the learned policy. Both \citet{emelianenko2019sequence} and \citet{wang2025learning} parameterise each generation step as position selection followed by value prediction, a choice that our ablations also find beneficial. More recent advances in generative modelling have also brought attention to \textbf{discrete diffusion models} and its generation order. We discuss relevant literature in Appendix~\ref{app:related_work}. While providing useful insights into order modelling, the data-based methods discussed above rely on target samples for training, whereas our focus is on sampling and order emergence under GFlowNet training without such samples. Additionally, we suggest a possible explanation for why order emerges, and our analysis method may also be applied to the models discussed above.


\section{Conclusion}

In this paper, we introduced three syntetic generation tasks: Markov-chain strings and string-based and graph-based checksum dependencies. 
We also formulated Sudoku as a sampling problem. On these tasks, we showed that the choice of backward policy and decomposition order affects sampler convergence. 
We proposed LO-GFlowNets, a training recipe that encourages the emergence of interpretable and useful generation orders without order annotations or problem-specific ordering rules. 
These findings suggest that LO-GFlowNets may uncover similar dependencies in new problems.
On checksum dependencies, our recipe converges more often than fixed random generation orders and the ablated learned backward policy configurations.
Finally, we proposed a possible explanation for the emergence of meaningful, non-random orders: orders associated with faster convergence exhibit higher gradient SNR at comparable loss values. 
This hypothesis suggests an operational criterion for a ``good'' order: one that provides a higher SNR under a matched evaluation protocol.

\textbf{Limitations.} LO-GFlowNets do not converge faster than naive autoregressive GFlowNets on the most complex among the considered tasks, Sudoku.
While the proposed method is useful for checksum dependencies and its extension to graphs, it is required to demonstrate a convergence advantage on more complex problems.
This would show a potential of LO-GFlowNets broad applicability. 
Our work should be extended with controlled problems where an optimal order is state-dependent, unlike Markov-chain strings and checksum dependencies, which exhibit ``global'' order independence from the state.

\bibliography{iclr2026_delta}
\bibliographystyle{iclr2026_delta}

\appendix

\clearpage
\section{Related work in discrete diffusion modelling}
\label{app:related_work}

In this section, we discuss discrete diffusion models and their adaptation to sampling from unnormalised distributions. Masked diffusion is closely related to any-order autoregressive modelling~\citep{hoogeboom2022autoregressive}. While considered autoregressive samplers fill one position per step, discrete diffusion samplers can update multiple positions per neural network evaluation. However, factorised parallel predictions can introduce approximation error by ignoring dependencies among simultaneously updated tokens. Both model families admit likelihood-based training, with diffusion models commonly trained through variational lower bounds.

\paragraph{Order in discrete diffusion models.}
\citet{kim2025train} show that standard masked-diffusion training averages over reveal orders and can encounter computationally difficult conditional prediction problems. Confidence-based unmasking can exploit more tractable conditionals at inference~\citep{chang2022maskgit,ye2025beyond,kim2025train}. A complementary approach is proposed by \citet{rout2026anchored}, who predict distributions over informative anchor tokens and condition denoising on these predictions, providing useful context without an explicit position-selection policy.

To learn order preferences during training, \citet{garg2025masked} optimise coordinate-specific masking schedules, which induce a state-independent distribution over reveal orders. \citet{bartosh2026forward} learn a data-dependent noising process to make few-step reverse generation easier. Addressing the training--inference mismatch, \citet{kim2026stop} construct training contexts by following an inference-time unmasking policy while revealing ground-truth tokens, improving training efficiency.

For variable-length generation, \citet{kim2026any} combine token insertion with unmasking while retaining flexibility in the unmasking order. Building on this formulation, \citet{patel2026insertion} jointly train the generator and data-dependent insertion and unmasking dynamics, demonstrating that generation order can also be learned through diffusion transition rates.

\paragraph{Discrete diffusion samplers.}
The discrete diffusion modelling paradigm also extends to sampling from unnormalised distributions without a dataset of target samples~\citep{sanokowski2025scalable,zhu2025mdns,guo2026proximal,pmlr-v306-carter26a}. \citet{sanokowski2025scalable} develop memory-efficient training objectives, while \citet{zhu2025mdns} formulate masked diffusion sampling through stochastic optimal control. \citet{guo2026proximal} introduce proximal updates on path measures to improve training stability and mode coverage, and \citet{pmlr-v306-carter26a} develop off-policy training strategies for discrete diffusion samplers. These methods share with GFlowNets a path-space formulation of sampling, but differ in their corruption kernels, transition structure, and training objectives. Our experiments focus on single-position construction actions. Extending our analysis to learned transition or unmasking schedules in discrete diffusion samplers is a promising direction for future work.

\clearpage
\section{Extended ablation study}
\label{app:ablation}

In this section, we revalidate our training recipe and repeat ablation study on Markov-chain strings. Additionally, we analyse learned generation orders by Learning-Order GFlowNet on Markov-chain strings and checksum dependencies.

\subsection{Markov-chain strings} 
Table~\ref{tab:markov_entropy_ablation} presents ablation study on Markov-chain strings. The results are essentially the same as on checksum dependencies, presented in the main section: our training recipe converges more often and faster than runs where some techniques are disabled.

\begin{table}[ht]
    \centering
    \caption{\textbf{Ablation study on Markov-chain strings}. Each configuration is launched with 5 random seeds. Average iterations include all seeds, assigning 100\,000 iterations to each non-converged run.}
    \label{tab:markov_entropy_ablation}
    \begin{tabular}{ll@{\hspace{1.5em}}rr}
        \toprule
        \multicolumn{2}{c}{Setup} & Converged & Avg. iterations \\
        \midrule
        w/o logits factorisation & $\lambda_{H_\text{B}}=1,\ \eta_{\mathrm{ord}}=0.1$ & 1/5 & 99\,100 \\
         & $\lambda_{H_\text{B}}=3,\ \eta_{\mathrm{ord}}=0.1$ & 0/5 & 100\,000 \\
        \midrule
        w/o backward entropy & $\lambda_{H_\text{B}}=0,\ \eta_{\mathrm{ord}}=0.1$ & 0/5 & 100\,000 \\
        \midrule
        w/o reduced order-head lr & $\lambda_{H_\text{B}}=1,\ \eta_{\mathrm{ord}}=1$ & 1/5 & 95\,500 \\
         & $\lambda_{H_\text{B}}=3,\ \eta_{\mathrm{ord}}=1$ & 1/5 & 98\,950 \\
        \midrule
        w/o uniform initialization & $\lambda_{H_\text{B}}=1,\ \eta_{\mathrm{ord}}=1$ & 3/5 & 81\,600 \\
         & $\lambda_{H_\text{B}}=1,\ \eta_{\mathrm{ord}}=0.1$ & 3/5 & 96\,300 \\
         & $\lambda_{H_\text{B}}=3,\ \eta_{\mathrm{ord}}=1$ & 0/5 & 100\,000 \\
         & $\lambda_{H_\text{B}}=3,\ \eta_{\mathrm{ord}}=0.1$ & 3/5 & 81\,650 \\
        \midrule
        & $\lambda_{H_\text{B}}=1,\ \eta_{\mathrm{ord}}=0.1$ & 4/5 & 79\,700 \\
        \textbf{Our training recipe} & $\lambda_{H_\text{B}}=3,\ \eta_{\mathrm{ord}}=0.1$ & 4/5 & \textbf{71\,850} \\
        & $\lambda_{H_\text{B}}=10,\ \eta_{\mathrm{ord}}=0.1$ & 4/5 & 75\,450 \\
        \hdashline
        with branching & $\lambda_{H_\text{B}}=3,\ \eta_{\mathrm{ord}}=0.1$ & \textbf{5/5} & 73\,450 \\
        \bottomrule
    \end{tabular}
\end{table}

\begin{figure}[p]
    \centering
    \includegraphics[width=\linewidth]{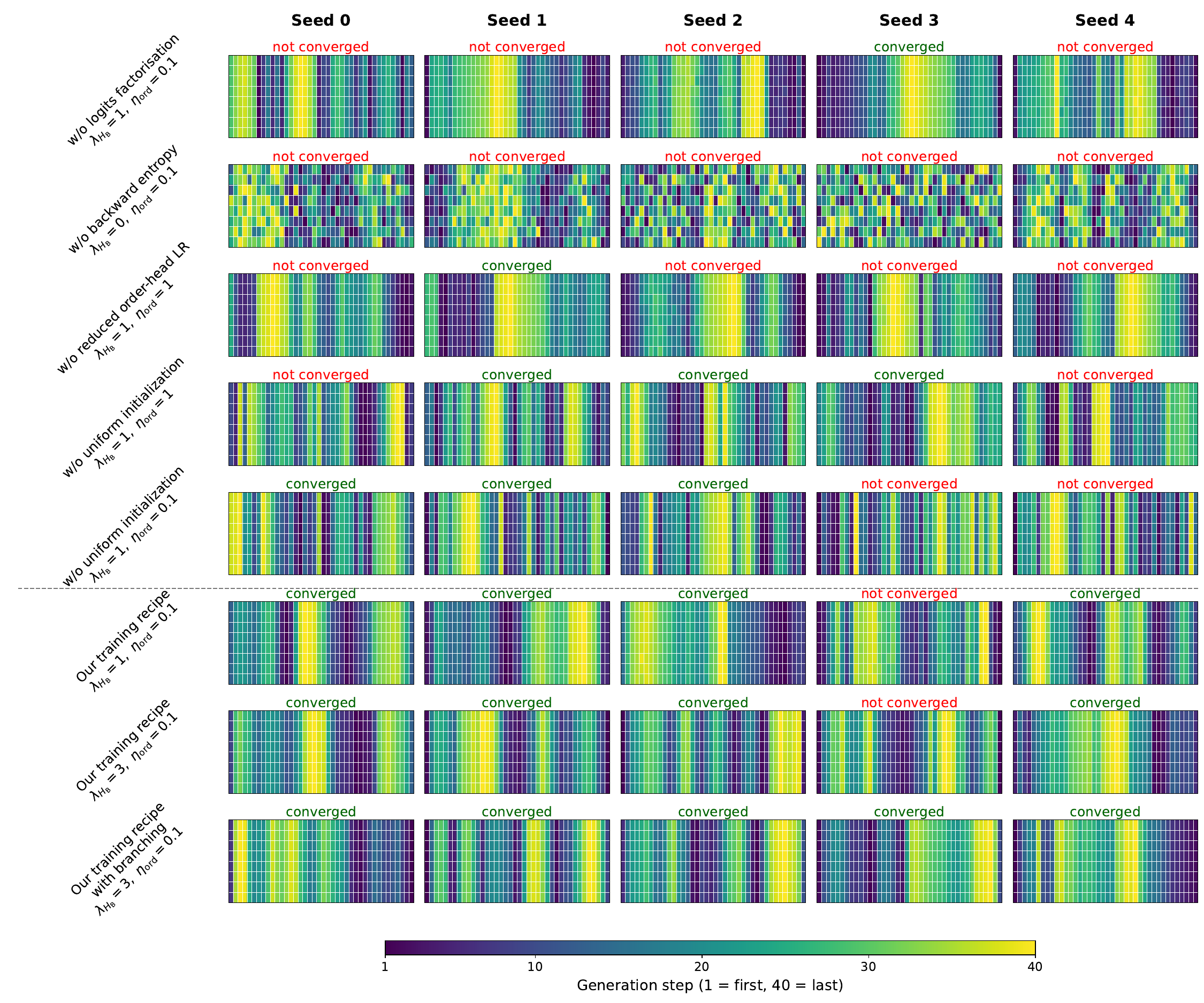}
    \caption{\textbf{Final generation orders for selected Markov-chain string ablations in Table~\ref{tab:markov_entropy_ablation}}. Each row is one configuration and each column is one seed. Within each panel, eight horizontal strips show independently sampled forward trajectories without exploration; the horizontal axis indexes the 40 string positions. Dark purple denotes the first generated position and yellow the last. Orders are sampled from each run's final available checkpoint. Panel titles indicate whether the run converged (dark green) or did not converge (red). The horizontal dashed line separates the ablated configurations from our training recipe and its variants.}
    \label{fig:markov_ablation_orders}
\end{figure}

Figure~\ref{fig:markov_ablation_orders} shows final generation orders for the selected Markov-chain string ablations. Overall, the samplers tend to generate strings block by block, choosing positions close to the previously generated symbol. Table~\ref{tab:markov_order_statistics} quantifies this tendency: the average distance between consecutive generated positions is generally smaller than under a uniformly random permutation. However, this statistic does not clearly distinguish our full training recipe from the ablated variants.

\begin{figure}[p]
    \centering
    \includegraphics[width=\linewidth,height=0.80\textheight,keepaspectratio]{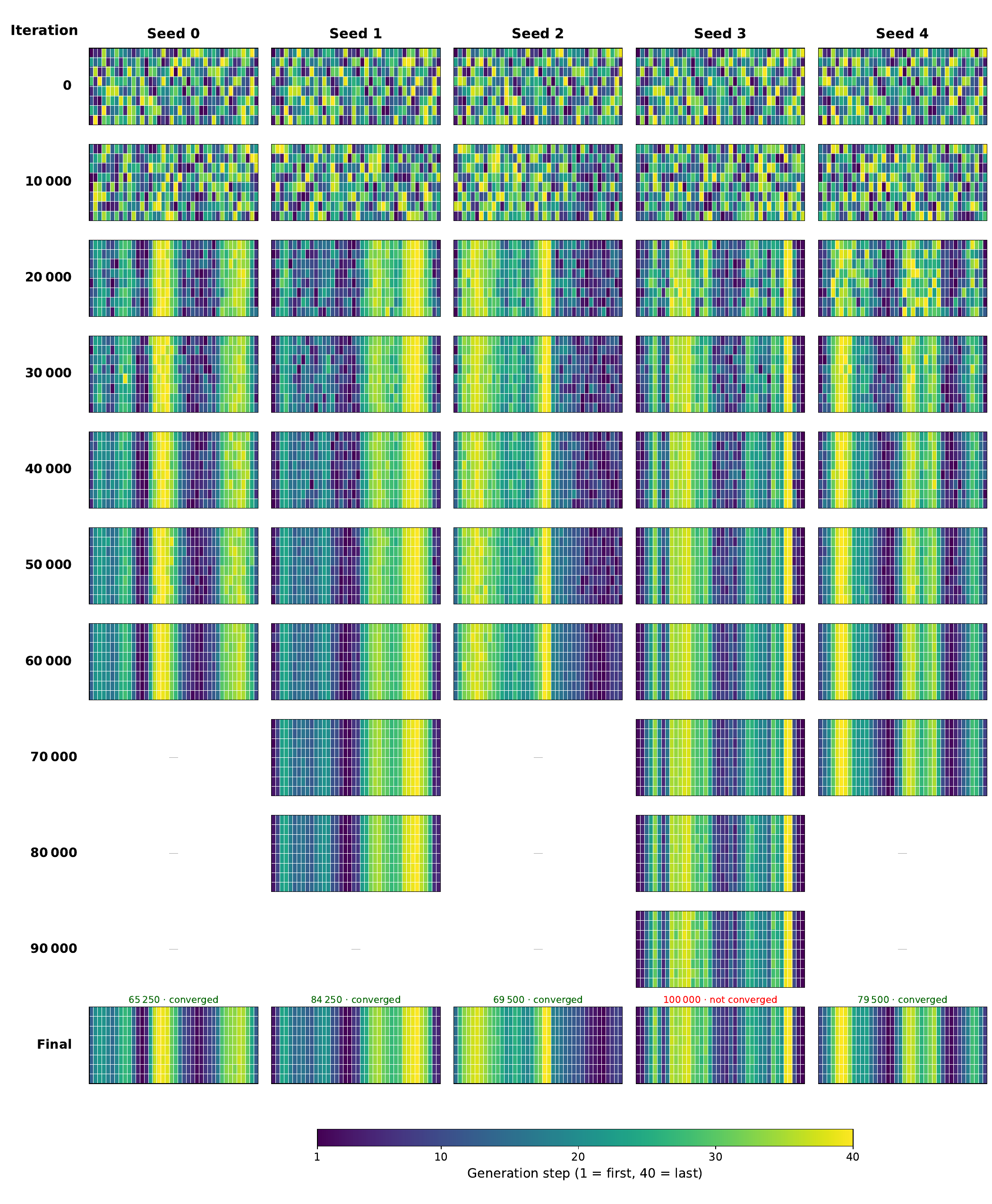}
    \caption{\textbf{Evolution of generation orders for our training recipe on Markov-chain strings}, with $\lambda_{H_\text{B}}=1$, uniform initialization, and $\eta_{\mathrm{ord}}=0.1$. Columns correspond to seeds and rows to training iterations, shown at initialization and every 10\,000 iterations. For seeds 0--2, intermediate snapshots are one iteration after the row labels. The bottom row shows each run's final checkpoint, with its exact iteration and convergence status. Each panel contains eight forward trajectories sampled without exploration across the 40 string positions. Dark purple denotes earlier generation and yellow later generation. Dashes indicate checkpoints beyond a run's stopping iteration.}
    \label{fig:markov_order_evolution}
\end{figure}

Figure~\ref{fig:markov_order_evolution} shows how generation orders evolve under our training recipe with $\lambda_{H_\text{B}}=1$. Training appears to involve an initial order-search phase, followed by improved symbol prediction under a largely stable order: the same pattern is seen on LO-GFlowNet on checksum dependencies, which we discuss later in this section. Block-wise patterns emerge within roughly the first $20\,000$--$40\,000$ iterations, well before convergence. 

\begin{table}[ht]
    \centering
    \caption{\textbf{Average distance between consecutive generated positions for the Markov-chain string ablations in Figure~\ref{fig:markov_ablation_orders}}. For a generation order $\pi$, we compute $\frac{1}{39}\sum_{i=1}^{39}|\pi_{i+1}-\pi_i|$ and average over five seeds and eight forward trajectories per seed, sampled without exploration, including non-converged runs. Smaller values indicate more local generation orders. Reference rows give the expected distance for a uniformly random permutation, $(40+1)/3$, and the distance for the identity order.}
    \label{tab:markov_order_statistics}
    \begin{tabular}{ll@{\hspace{1.5em}}r}
        \toprule
        \multicolumn{2}{c}{Setup} & Average distance \\
        \midrule
        w/o logits factorisation & $\lambda_{H_\text{B}}=1,\ \eta_{\mathrm{ord}}=0.1$ & 8.26 \\
        \midrule
        w/o backward entropy & $\lambda_{H_\text{B}}=0,\ \eta_{\mathrm{ord}}=0.1$ & 13.36 \\
        \midrule
        w/o reduced order-head lr & $\lambda_{H_\text{B}}=1,\ \eta_{\mathrm{ord}}=1$ & 10.18 \\
        \midrule
        w/o uniform initialization & $\lambda_{H_\text{B}}=1,\ \eta_{\mathrm{ord}}=1$ & 11.90 \\
         & $\lambda_{H_\text{B}}=1,\ \eta_{\mathrm{ord}}=0.1$ & 10.97 \\
        \midrule
        \textbf{Our training recipe} & $\lambda_{H_\text{B}}=1,\ \eta_{\mathrm{ord}}=0.1$ & 9.49 \\
         & $\lambda_{H_\text{B}}=3,\ \eta_{\mathrm{ord}}=0.1$ & 11.39 \\
        \hdashline
        with branching & $\lambda_{H_\text{B}}=3,\ \eta_{\mathrm{ord}}=0.1$ & 9.03 \\
        \midrule
        Uniformly random order & Reference & 13.67 \\
        Identity order & Reference & 1.00 \\
        \bottomrule
    \end{tabular}
\end{table}

\newpage
\subsection{Checksum dependencies}
\begin{figure}[p]
    \centering
    \includegraphics[width=\linewidth]{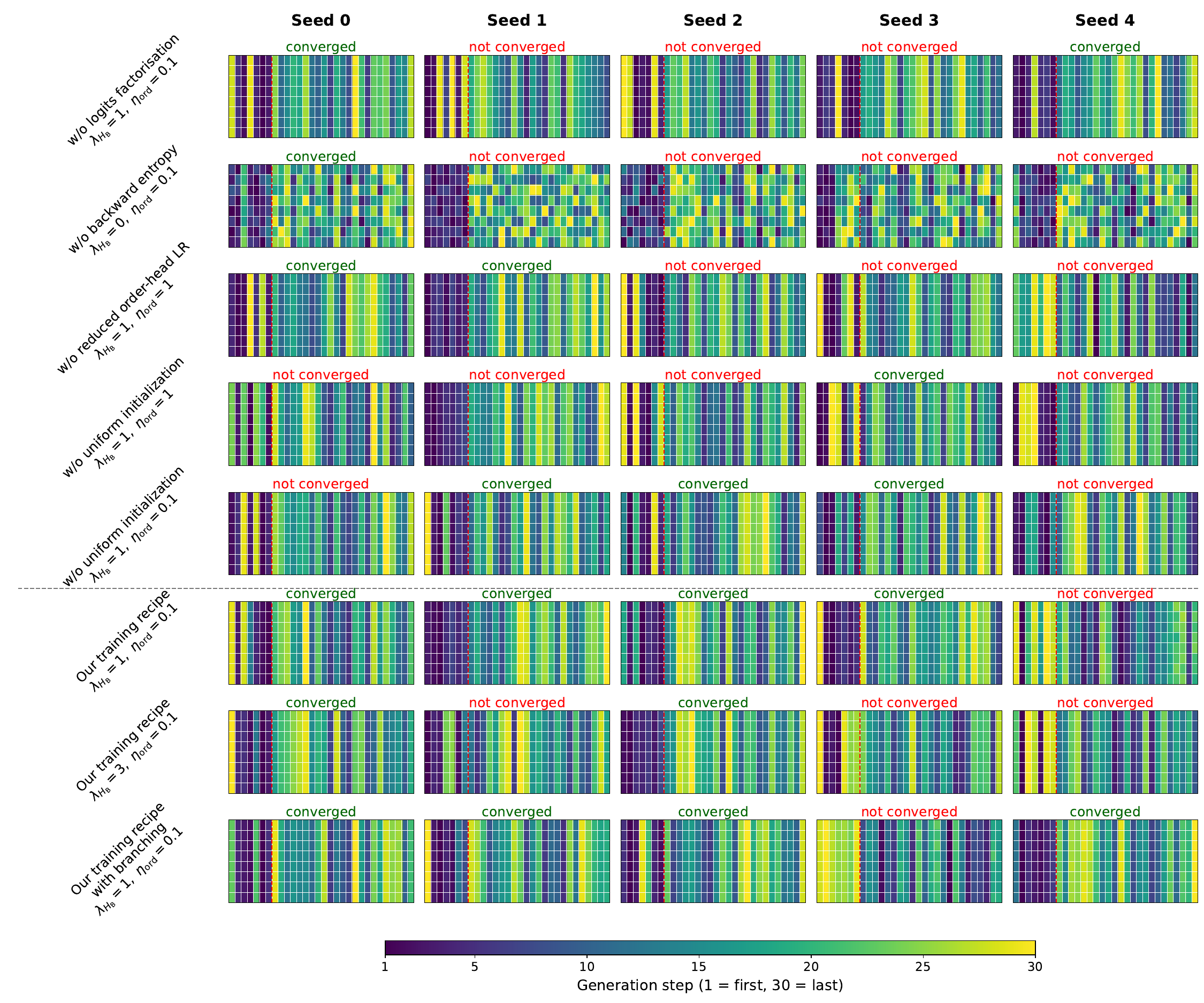}
    \caption{\textbf{Final generation orders for the checksum dependencies ablations in Table~\ref{tab:checksum_entropy_ablation}}. Each row is one configuration and each column is one seed. Within each panel, eight horizontal strips show independently sampled forward trajectories without exploration; the horizontal axis indexes the 30 string positions. Dark purple denotes the first generated position and yellow the last. The red dashed line separates the seven-position prefix from the suffix. Panel titles indicate whether the run converged (dark green) or did not converge (red); non-converged runs reached the 100\,000-iteration limit. The horizontal dashed line separates the ablated configurations from our training recipe and its variants.}
    \label{fig:checksum_ablation_orders}
\end{figure}

Figure~\ref{fig:checksum_ablation_orders} shows generation orders sampled from the final checkpoint of each checksum ablation run: darker colors reflect earlier generation time. Overall, the converged runs tend to generate the prefix earlier.
A configuration without logits factorisation sometimes learn prefix-first order, but more often than our training recipe.
A configuration with disable backward entropy also tends to generate the prefix first, but it still does not converge within the allocated training budget. Our training recipe tends to follow prefix-order more often than configurations without uniform initialisation and reduced order-head learning rate.

Table~\ref{tab:checksum_order_statistics} quantifies the average generation step of prefix and suffix symbols for these trajectories, quantitavely showing that our training recipe leads to prefix-early generation order.

\begin{figure}[p]
    \centering
    \includegraphics[width=\linewidth,height=0.80\textheight,keepaspectratio]{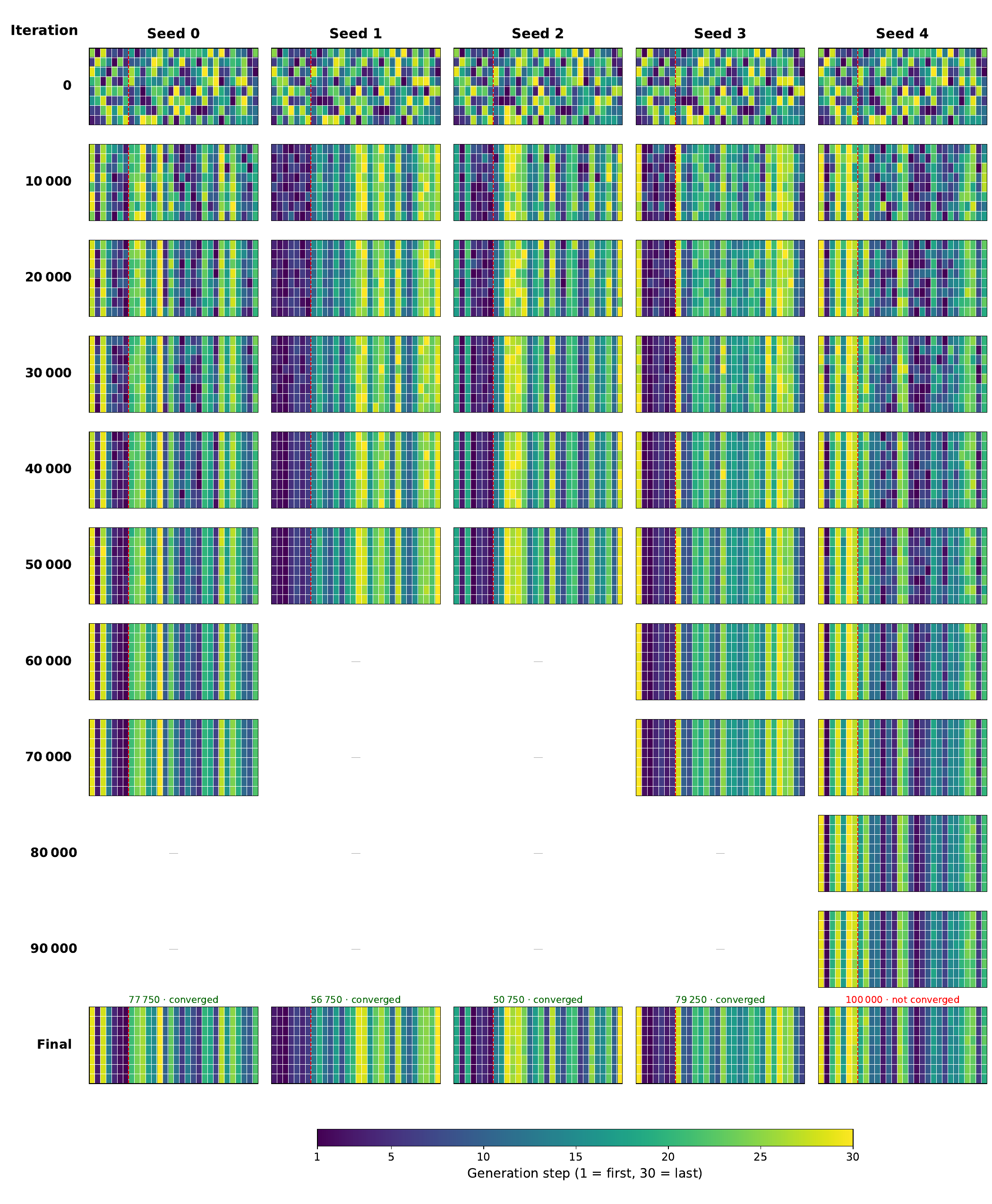}
    \caption{\textbf{Evolution of generation orders for our training recipe on checksum dependencies}, with $\lambda_{H_\text{B}}=1$, uniform initialization, and $\eta_{\mathrm{ord}}=0.1$. Columns correspond to seeds and rows to training iterations, shown at initialization and every 10\,000 iterations. The bottom row shows each run's final checkpoint, with its iteration and convergence status. Each panel contains eight forward trajectories sampled without exploration. Dark purple denotes earlier generation and yellow later generation; the red dashed line separates the seven-position prefix from the suffix. Dashes indicate checkpoints beyond a run's stopping iteration.}
    \label{fig:checksum_order_evolution}
\end{figure}

Figure~\ref{fig:checksum_order_evolution} shows the evolution of generation orders under our training recipe with $\lambda_{H_\text{B}}=1$ across random seeds. As on Markov-chain strings, training appears to follow two phases: order search during roughly the first $40\,000$ iterations, followed by improved symbol prediction under a largely stable order until convergence: LO-GFlowNet initially exhibits convergence speed close to that of the GFlowNet with the uniform backward policy. Another interesting observation is that seeds 0 and 3 converge later than seeds 1 and 2. The orders of the former retain bright, late-generated positions within the prefix (Figure~\ref{fig:checksum_order_evolution}), suggesting that these orders are less favourable than those learned by seeds 1 and 2.

\begin{table}[ht]
    \centering
    \caption{\textbf{Average generation step for prefix and suffix symbols for the checksum dependencies ablations in Table~\ref{tab:checksum_entropy_ablation}}. Steps are numbered from 1 to 30. Each configuration averages over five seeds and eight forward trajectories per seed, sampled without exploration, including non-converged runs. Prefix positions are 1--7 and suffix positions are 8--30. The last column is the suffix average minus the prefix average; larger values indicate a stronger average preference for generating prefix symbols earlier. Reference rows give the expected averages under uniformly random orders and the averages when all prefix symbols precede all suffix symbols.}
    \label{tab:checksum_order_statistics}
    \begin{tabular}{ll@{\hspace{1.5em}}rrr}
        \toprule
        \multicolumn{2}{c}{Setup} & Prefix avg. & Suffix avg. & Difference \\
        \midrule
        w/o logits factorisation & $\lambda_{H_\text{B}}=1,\ \eta_{\mathrm{ord}}=0.1$ & 10.52 & 17.02 & 6.50 \\
        \midrule
        w/o backward entropy & $\lambda_{H_\text{B}}=0,\ \eta_{\mathrm{ord}}=0.1$ & 7.68 & 17.88 & 10.21 \\
        \midrule
        w/o reduced order-head lr & $\lambda_{H_\text{B}}=1,\ \eta_{\mathrm{ord}}=1$ & 12.51 & 16.41 & 3.89 \\
        \midrule
        w/o uniform initialization & $\lambda_{H_\text{B}}=1,\ \eta_{\mathrm{ord}}=1$ & 12.40 & 16.44 & 4.04 \\
         & $\lambda_{H_\text{B}}=1,\ \eta_{\mathrm{ord}}=0.1$ & 9.49 & 17.33 & 7.84 \\
        \midrule
        \textbf{Our training recipe} & $\lambda_{H_\text{B}}=1,\ \eta_{\mathrm{ord}}=0.1$ & 10.36 & 17.06 & 6.70 \\
         & $\lambda_{H_\text{B}}=3,\ \eta_{\mathrm{ord}}=0.1$ & 11.80 & 16.63 & 4.83 \\
        \hdashline
        with branching & $\lambda_{H_\text{B}}=1,\ \eta_{\mathrm{ord}}=0.1$ & 11.66 & 16.67 & 5.01 \\
        \midrule
        Uniformly random order & Reference & 15.50 & 15.50 & 0.00 \\
        All prefix symbols first & Reference & 4.00 & 19.00 & 15.00 \\
        \bottomrule
    \end{tabular}
\end{table}

\clearpage
\section{Experiments}
\label{app:experiments}

\subsection{Markov-chain strings}

\begin{figure}[ht]
\centering
    \caption{\textbf{Results on Markov-chain strings.} \textit{Left}: number of converged runs and average number of iterations, assigning 100\,000 iterations to non-converged runs. Fixed orders with $K$ random swaps, random fixed orders, and uniform $P_B$ use ten seeds each. LO-GFlowNet uses our training recipe with branching, $\lambda_{H_\text{B}}=3$, and $\eta_{\mathrm{ord}}=0.1$, evaluated over five seeds (Table~\ref{tab:markov_entropy_ablation}). \textit{Right}: ELBO (top) and EUBO (bottom) for $K\in\{0,40\}$, uniform $P_B$, and LO-GFlowNet. Curves show the 50th percentile and shaded bands the 20th--80th percentiles, computed in base-10 log-space for $-\mathrm{ELBO}$ and $\mathrm{EUBO}$, with the ELBO sign restored. Converged runs retain their final metrics at later iterations; dashed black lines mark the convergence thresholds.}
    \label{fig:markov_chain_convergence}
    \begin{minipage}[c]{0.39\linewidth}
    \centering
    \small
    \setlength{\tabcolsep}{3pt}
    \begin{tabular}{lcc}
    \toprule
    Sampler & Converged & \shortstack{Avg.\\iterations} \\
    \midrule
    \shortstack[l]{Fixed order\\$K=0$} & \textbf{10/10} & $\mathbf{7\,275}$ \\
    $K=5$ & \textbf{10/10} & $21\,275$ \\
    $K=10$ & \textbf{10/10} & $29\,250$ \\
    $K=20$ & \textbf{10/10} & $37\,475$ \\
    $K=40$ & \textbf{10/10} & $50\,600$ \\
    \hdashline
    \shortstack[l]{Random\\fixed order} & \textbf{10/10} & $42\,825$ \\
    \hdashline
    Uniform $P_B$ & 0/10 & $100\,000$ \\
    \midrule
    LO-GFlowNet & \textbf{5/5} & $73\,450$ \\
    \bottomrule
    \end{tabular}
    \end{minipage}\hfill
    \begin{minipage}[c]{0.60\linewidth}
    \centering
    \includegraphics[width=\linewidth]{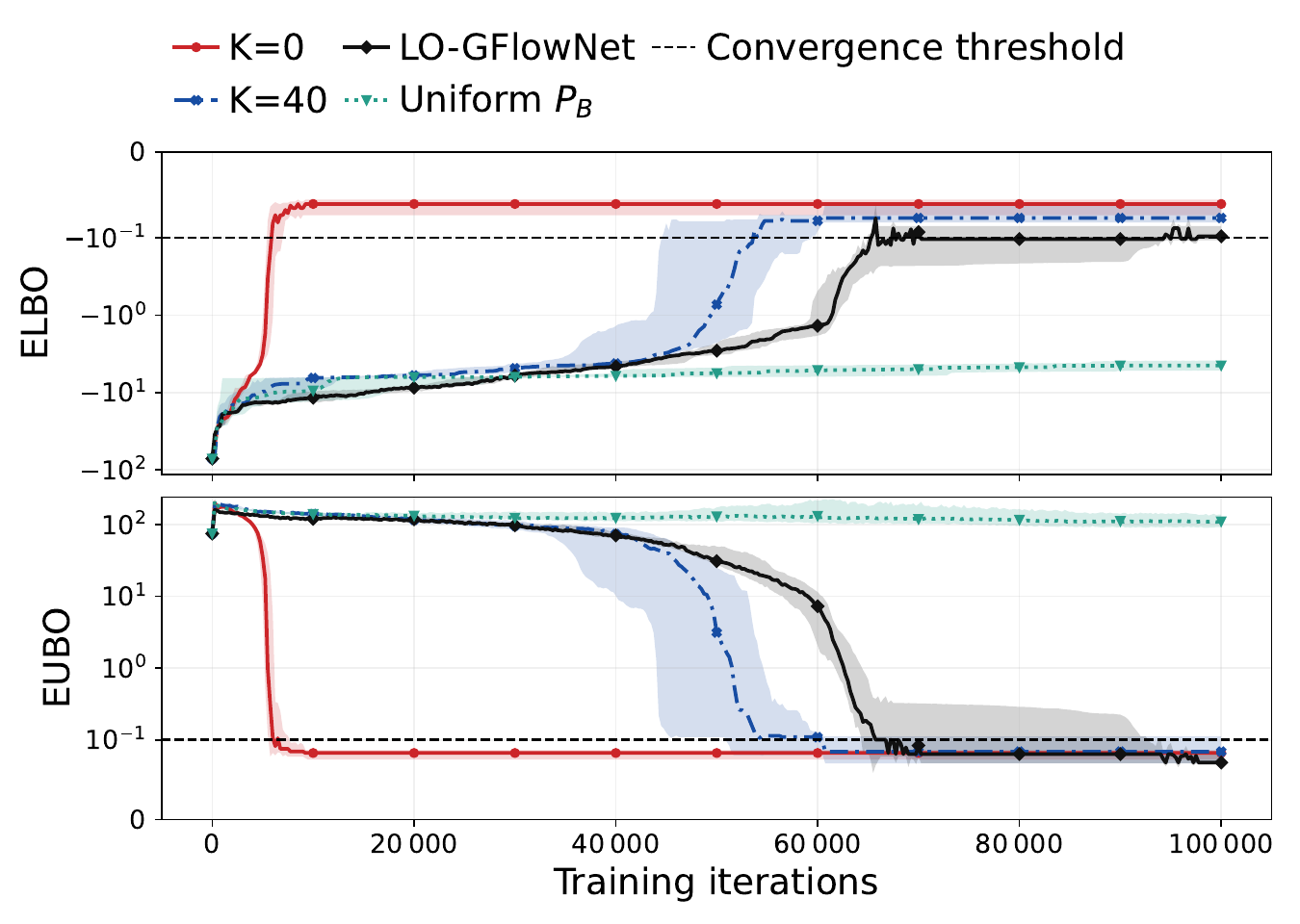}
    \end{minipage}
\end{figure}

As for checksum dependencies, we also consider samplers with fixed generation orders during training. To construct a progression of less suitable orders, define $\pi^{\text{MC}}_0$ as the identity permutation and $\pi^{\text{MC}}_K$ as a permutation with applied $K$ random swaps. 
\figref{fig:markov_chain_convergence} (left) shows that average number of training iterations to convergence increases with $K$. \figref{fig:markov_chain_convergence} (right) compares ELBO and EUBO for $K\in\{0,40\}$, uniform $P_B$, and LO-GFlowNet. As with checksum dependencies, higher $K$ leads to later convergence. However, any order can converge in Markov-chain strings, and only uniform $P_B$ is suboptimal. LO-GFlowNet starts with the uniform backward policy and spends initial budget to find a suitable order as shown in Figure~\ref{fig:markov_order_evolution}, hence, it converges later than the GFlowNet with $\pi^{\text{MC}}_{40}$-order.

\subsection{Sudoku}
\label{app:sudoku_experiments}

We study conditional Sudoku generation at two simplification levels, $p=10\%$ and $p=30\%$, where $p$ is the fraction of originally empty cells revealed before training and evaluation. Each puzzle defines a separate sampling problem, and its clues remain fixed. We compare raster-order autoregression, autoregression with a uniformly sampled fixed random order, solver-guided autoregression, and learned generation orders. As a solver-guided order, we use TDoku~\citep{dillon2019tdoku}, an efficient non-neural Sudoku solver combining constraint propagation with search. We recompute its cell choice on the current board; the neural network still predicts the digit. We report correct-digit accuracy, ELBO, and dataset-based IW-EUBO (Appendix~\ref{app:metrics}) on the first $50\,000$ held-out puzzles. Correct-digit accuracy uses greedy digit prediction, whereas ELBO and IW-EUBO use categorical digit sampling.

\begin{figure}[ht]
    \centering
    \includegraphics[width=\linewidth]{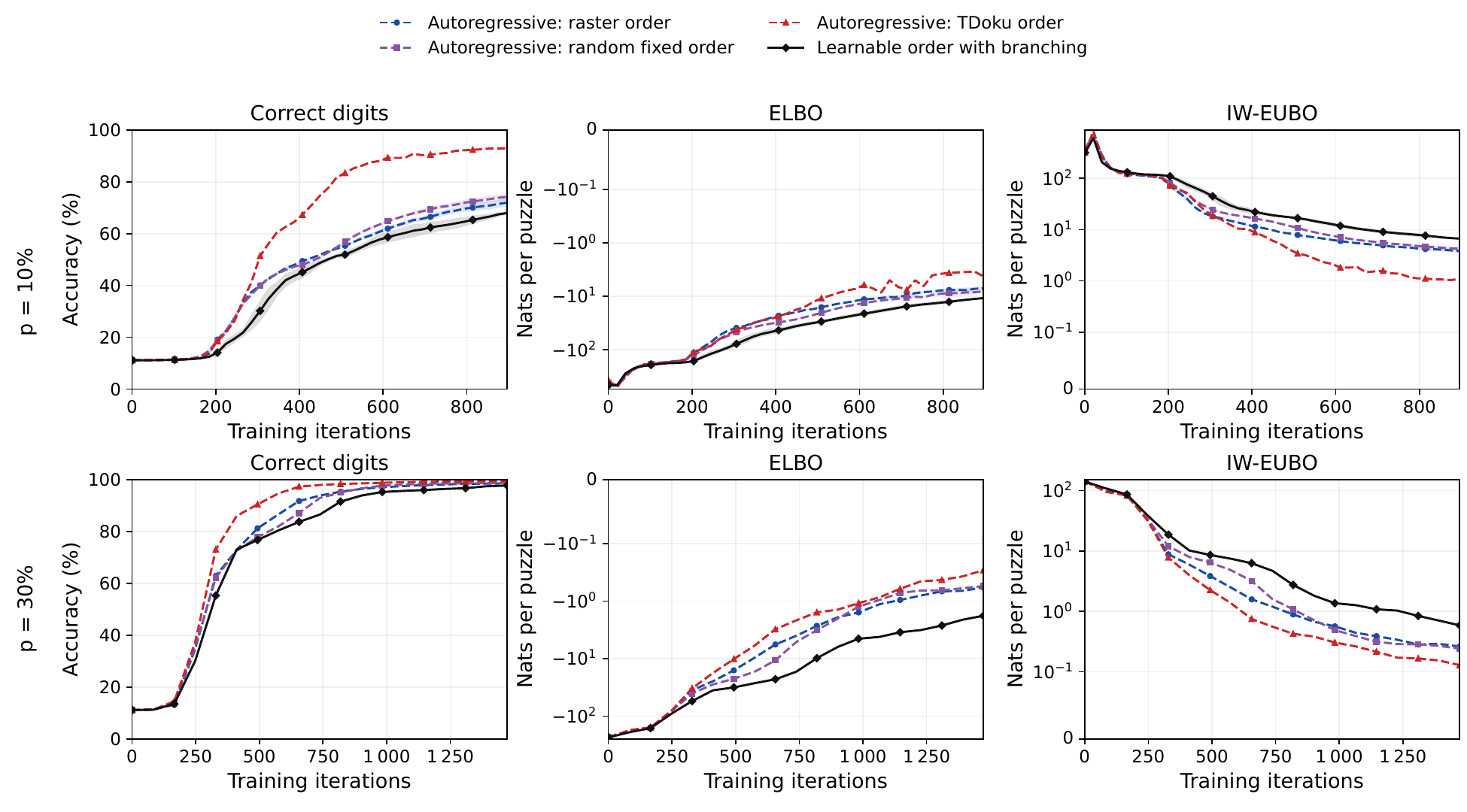}
    \caption{\textbf{Sudoku with ordinary decoding.} Correct digits, ELBO, and dataset-based IW-EUBO for $p=10\%$ (top) and $p=30\%$ (bottom), using each model's original generation-order policy. For $p=10\%$, curves show means and bands show one standard deviation across three seeds, except TDoku, which has one seed; statistics use shared evaluation iterations within each configuration. For $p=30\%$, each configuration has one seed. For each difficulty, all curves are truncated at the latest evaluation shared by the configurations and seed cohorts: iteration 896 for $p=10\%$ and $1\,473$ for $p=30\%$, without interpolation or extrapolation. ELBO and IW-EUBO axes use a symmetric logarithmic scale with a linear region around zero.}
    \label{fig:sudoku_elbo_eubo}
    \label{fig:sudoku_p30_comparison}
\end{figure}

\paragraph{Order in Sudoku matters: TDoku-order GFlowNets converge faster.}
Figure~\ref{fig:sudoku_elbo_eubo} compares generation-order strategies on both difficulty levels. TDoku-guided generation achieves higher correct-digit accuracy and improves ELBO and IW-EUBO earlier in training than raster and fixed random orders. For example, on $p=30\%$ at iteration 901, TDoku order achieves $98.60\%$ correct digits, compared with $96.46\%$ for raster order and $96.70\%$ for fixed random order. These results show that the choice of generation order affects convergence in Sudoku and that a solver-guided order can accelerate training.

\begin{figure}[ht]
    \centering
    \includegraphics[width=\linewidth]{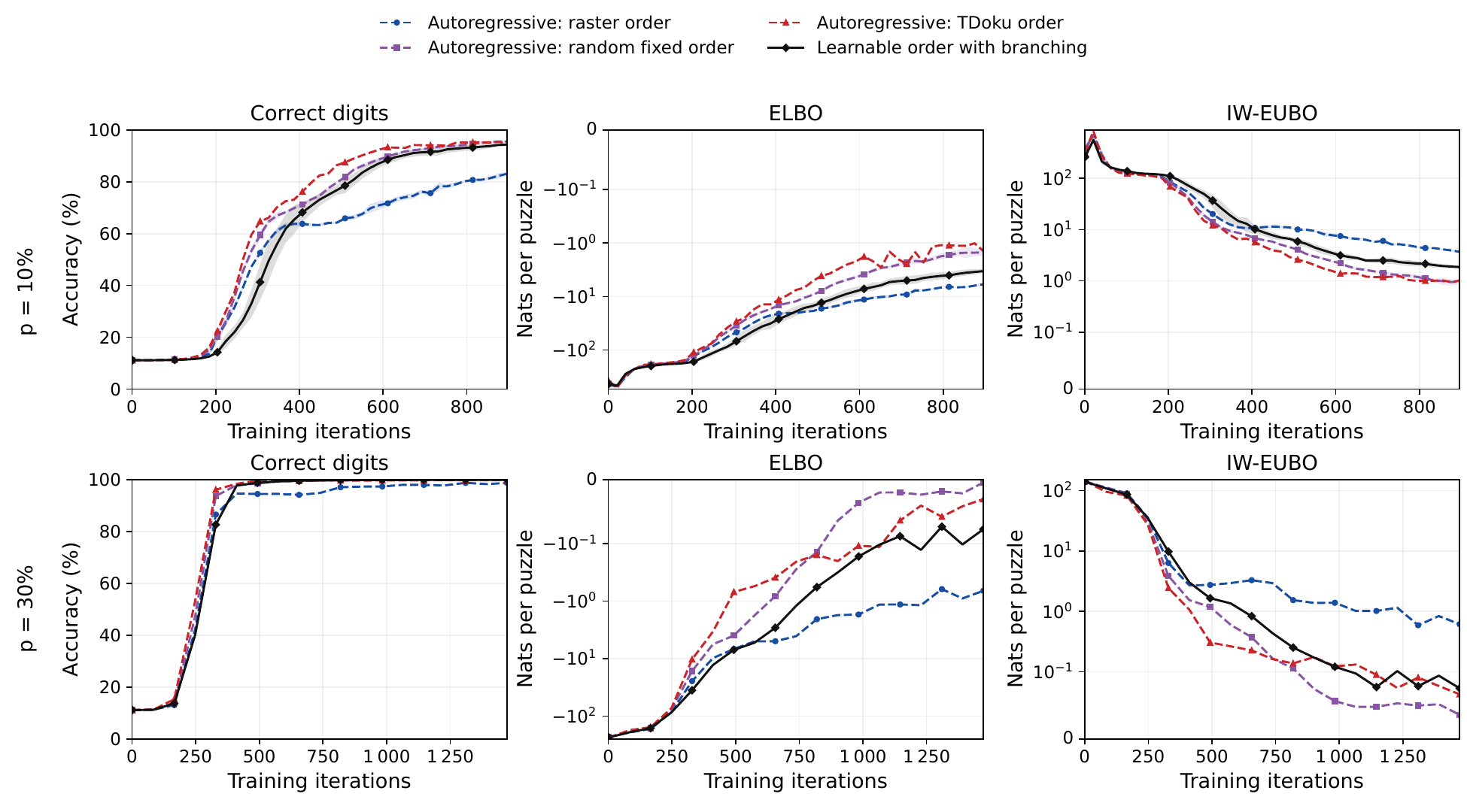}
    \caption{\textbf{Sudoku with confidence-based ordering.} The same models and metrics as in Figure~\ref{fig:sudoku_elbo_eubo}, evaluated by selecting the empty cell whose most probable digit has the highest predicted probability. Rows correspond to $p=10\%$ (top) and $p=30\%$ (bottom). Correct-digit accuracy uses greedy digit prediction; ELBO and IW-EUBO use categorical digit sampling. Models are not retrained. Seed cohorts, averaging, and axis scales match Figure~\ref{fig:sudoku_elbo_eubo}.}
    \label{fig:sudoku_confidence_comparison}
\end{figure}

\paragraph{Confidence sampling improves performance, but learning a useful generation order remains challenging.}
The learned-order model uses branching but does not achieve the same convergence advantage as TDoku under its own position policy. To assess whether its digit predictions support more effective orders, we evaluate confidence-based ordering: at each step, it chooses the empty cell whose most probable digit has the highest predicted probability. This changes only the evaluation-time position choice and does not retrain the model. Figure~\ref{fig:sudoku_confidence_comparison} shows that this rule substantially improves the learned-order model's correct-digit accuracy and sampling metrics at both difficulty levels. A model trained with a fixed random order also benefits from this intervention. Thus, the digit network can support better generation orders than those selected by the learned position policy, but learning to exploit this information remains a limitation of our approach.

\begin{figure}[p]
    \centering
    \includegraphics[width=\linewidth,height=0.82\textheight,keepaspectratio]{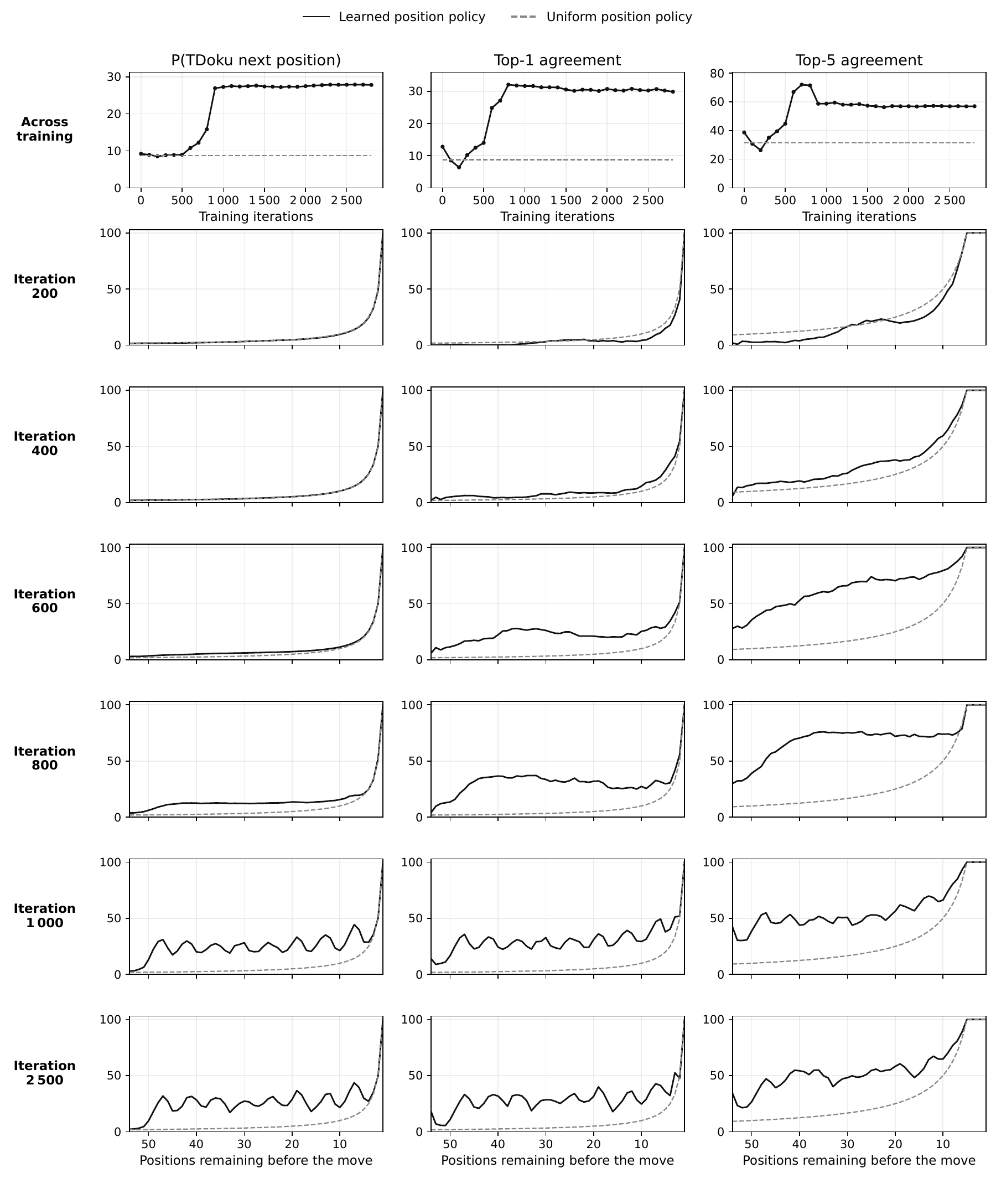}
    \caption{\textbf{TDoku alignment while filling Sudoku puzzles, $p=10\%$.} The top row shows aggregate alignment over training, evaluated every 100 iterations. The remaining rows show training checkpoints at iterations 200, 400, 600, 800, $1\,000$, and $2\,500$. Columns show the probability assigned to TDoku's next position (left), top-1 agreement (middle), and top-5 agreement (right), all in percent. In the top row, the horizontal axis is training iteration; in the remaining rows, it is the actual number of positions remaining before the move, decreasing from left to right. Each checkpoint uses $1\,024$ fixed held-out puzzles; positions are sampled from the model, correct reference digits are inserted, and TDoku is recomputed at every move. Dashed curves show uniform-position references, $1/r$ for probability and top-1, and $\min(5/r,1)$ for top-5, where $r$ is the number of remaining positions. Results are averaged over states with the same $r$, without smoothing. This separate diagnostic run differs from the three-seed $p=10\%$ quality comparison.}
    \label{fig:sudoku_tdoku_alignment}
\end{figure}

\paragraph{Alignment with a solver order.}
In a separate $p=10\%$ diagnostic run, we evaluate agreement with TDoku on $1\,024$ fixed held-out puzzles. At every step, we sample a position from the model, insert its correct reference digit, and recompute the solver's next position. Agreement is scored using the model's original position probabilities, without confidence-based ordering. Figure~\ref{fig:sudoku_tdoku_alignment} shows the probability assigned to TDoku's next position and top-1 and top-5 agreement as puzzles are filled, at selected training checkpoints. The top row summarises alignment over training, while the remaining rows group measurements by the actual number of empty positions remaining before each move. The drop in top-5 agreement and the cosine-like oscillations that appear around iteration $1\,000$ reflect the sampler settling into $3 \times 3$ block-by-block generation. It continues to follow this block order throughout the remaining observed training period. The learned policy can therefore exhibit nontrivial solver alignment without reproducing the solver's order. This diagnostic uses correct-digit trajectories and does not measure behaviour after the model makes a digit error.

\clearpage
\section{Additional results on gradients SNR}
\label{app:snr}

\paragraph{Markov-chain strings.}

\begin{figure}[p]
    \centering
    \includegraphics[width=\linewidth]{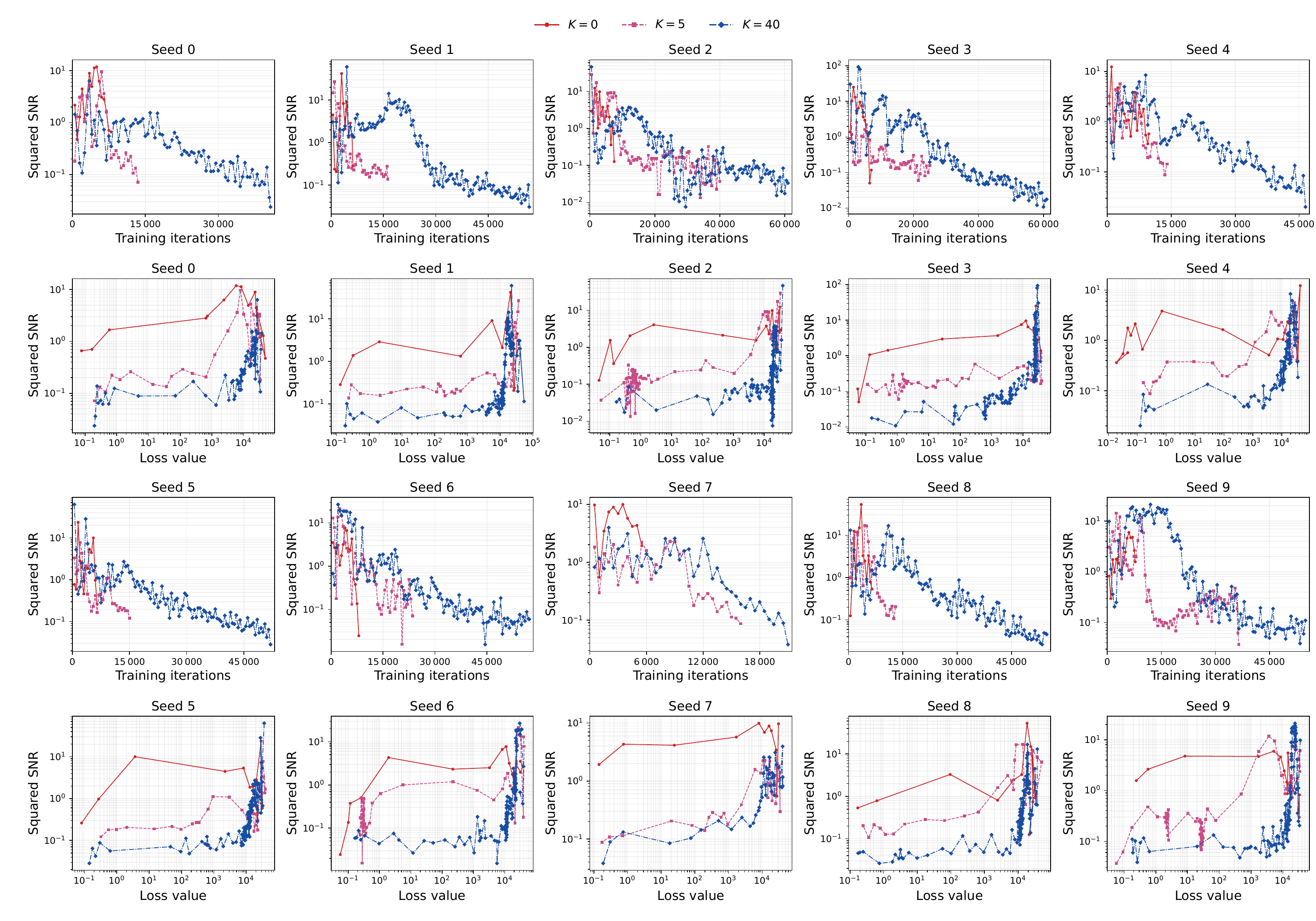}
    \caption{\textbf{Markov-chain strings: squared SNR on current-policy trajectories.} Gradients are measured on pure forward-policy samples; the loss axis retains the original target-based full-dataset TB loss. \textit{The first two rows} show seeds 0--4 and \textit{the last two} show seeds 5--9. Within each block, \textit{the upper row} plots squared SNR against training iterations and \textit{the lower row} against validation loss. All checkpoints after initialization are connected in training order, without smoothing or regression fits.}
    \label{fig:markov_snr_all_seeds}
\end{figure}

Figure~\ref{fig:markov_snr_all_seeds} extends the Markov-chain string comparison in Figure~\ref{fig:markov_snr_main} to all ten seeds for $K\in\{0,5,40\}$, showing squared SNR over training iterations and against validation loss. At each checkpoint, the reference gradient is computed from $131\,072$ fresh trajectories sampled from the current forward policy without exploration. Noise is estimated using 128 minibatches of 256 trajectories, independently sampled with replacement from this trajectory dataset. The validation loss is evaluated separately on a fixed dataset of exact-target terminals, with trajectories constructed using the current backward policy. Thus, SNR describes gradients on the model's own trajectories, while the loss axis measures progress against the target distribution.

\paragraph{Checksum dependencies.}

\begin{figure}[p]
    \centering
    \includegraphics[width=\linewidth]{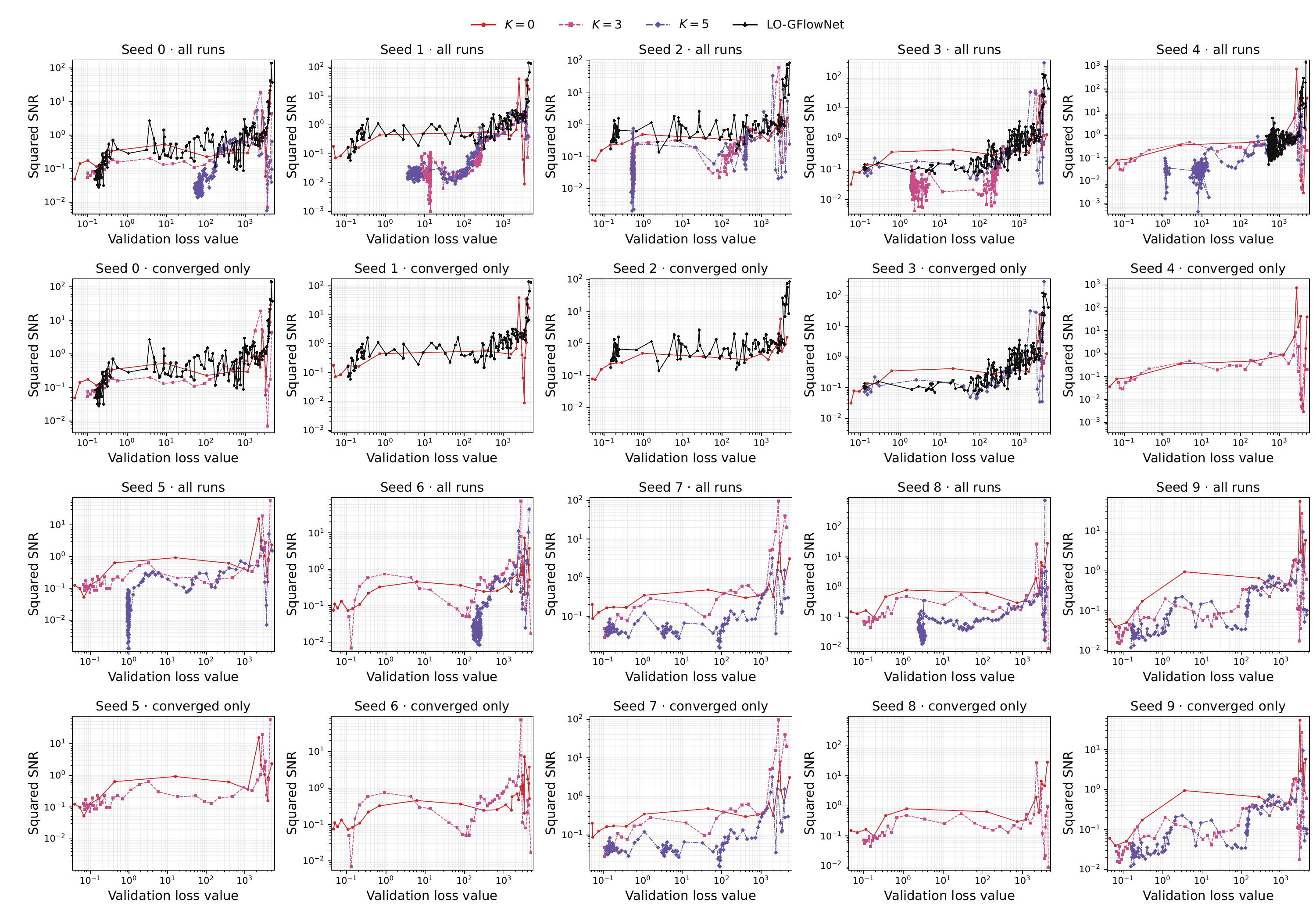}
    \caption{\textbf{Checksum dependencies: squared SNR versus validation loss, with and without convergence-based selection.} Gradients are evaluated on current-policy trajectories, while the horizontal axis uses the separate target-based validation TB loss. \textit{The first two rows} show seeds 0--4 and \textit{the last two} show seeds 5--9. Within each block, \textit{the upper row} includes all runs and \textit{the lower row} includes only converged runs. Each seed's paired panels use the same axis limits. LO-GFlowNet was trained on seeds 0--4 only. All valid checkpoints after initialization are connected in training order, without smoothing, loss-based filtering, or regression fits.}
    \label{fig:checksum_snr_all_seeds}
\end{figure}

Figure~\ref{fig:checksum_snr_all_seeds} shows squared SNR against validation loss for checksum dependencies with $K\in\{0,3,5\}$ and LO-GFlowNet. Within each two-row block, the upper row includes all runs and the lower row retains only converged runs, using matching axis limits for each seed. The converged subset contains ten runs for $K=0$, seven for $K=3$, three for $K=5$, and four for LO-GFlowNet. All valid checkpoints are shown; the loss threshold of $10^3$ applies only to the regression fits in Figure~\ref{fig:markov_snr_main}. LO-GFlowNet uses our training recipe without branching, with uniform initialization, $\eta_{\mathrm{ord}}=0.1$, and $\lambda_{H_\text{B}}=1$.

\clearpage
\section{Implementation details}
\label{app:implementation_details}

In all experiments, we use a jax-based library for GFlowNets, gfnx~\citep{tiapkin2025gfnx}, that enables fast experimentation.

\subsection{Environments}

\paragraph{General intuition.} One useful feature of the environments we consider is that their difficulty can be tuned. For Markov-chain strings and checksum dependencies, the length $L$ and alphabet size $A$ control the size of the state space, while the noise level $\epsilon$ controls the concentration of the target distribution: smaller $\epsilon$ makes it sharper. For Sudoku, the board size is fixed at $9 \times 9$, but the initial puzzles can be simplified. We construct easier variants by revealing the $p$-percent correct digits of the initially unfilled cells. As our experiments show, increasing $p$ leads to significantly faster convergence, yielding easier problems.

\begin{figure}[ht]
    \centering
    \includegraphics[width=\linewidth]{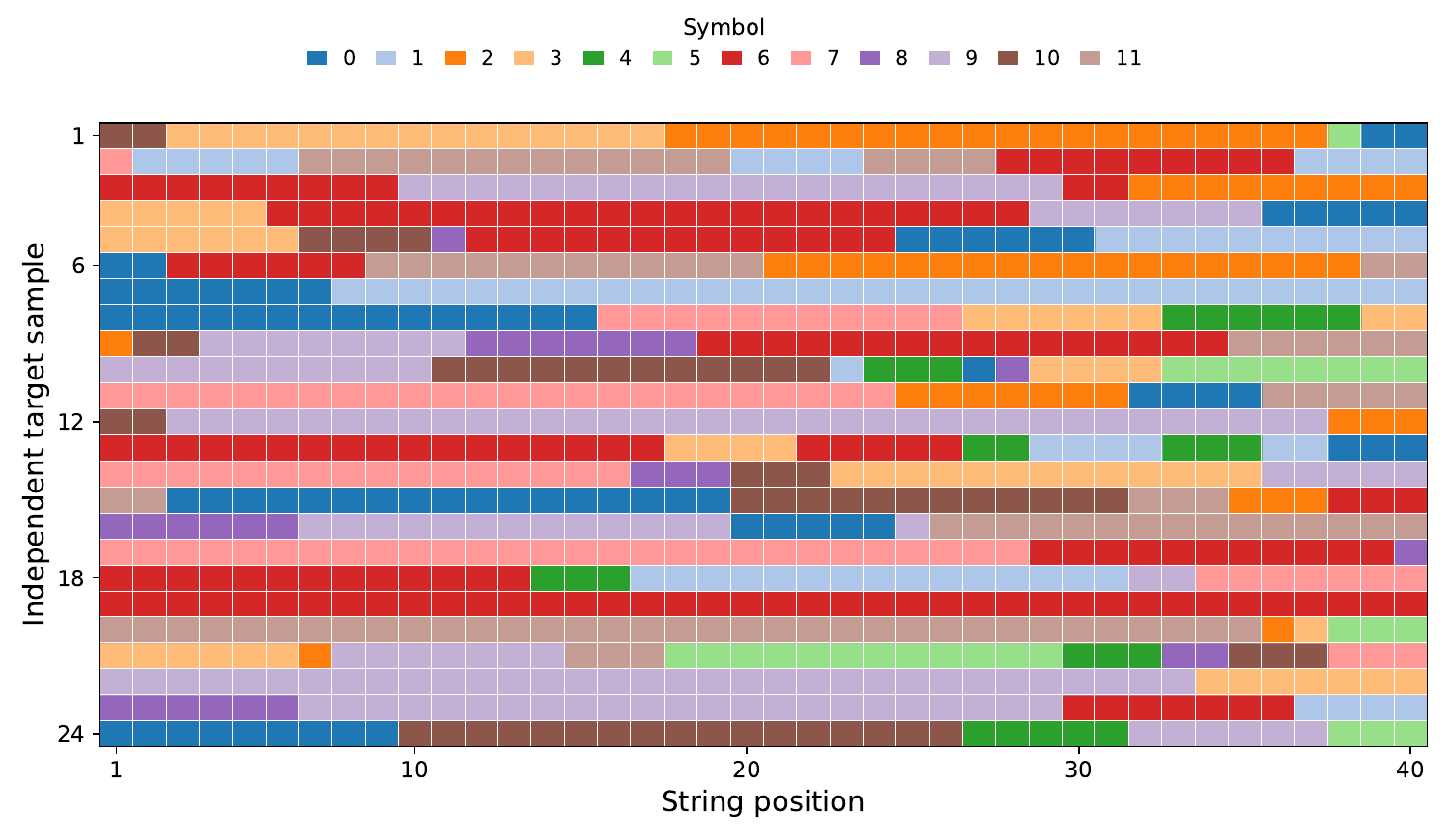}
    \caption{Exact target samples for Markov-chain strings with $L=40$, $A=12$, and $\epsilon=0.1$. Each row is an independently sampled string, and each colour represents a symbol. Long stretches of identical symbols reflect the probability $0.9$ of repeating the preceding symbol. The 24 samples are shown without selection or sorting.}
    \label{fig:markov_target_samples}
\end{figure}

\paragraph{Markov-chain strings.}
We use string length $L=40$, alphabet size $A=12$, and transition noise $\epsilon=0.1$. Thus, the first symbol is uniform over the 12 symbols, and each subsequent symbol repeats its predecessor with probability $0.9$; each of the other 11 symbols has probability $0.1/11$. The exact normalisation constant is $Z=1$. 
Figure~\ref{fig:markov_target_samples} shows the independent samples from the exact target distribution illustrating the local dependencies.


\begin{figure}[ht]
    \centering
    \includegraphics[width=\linewidth]{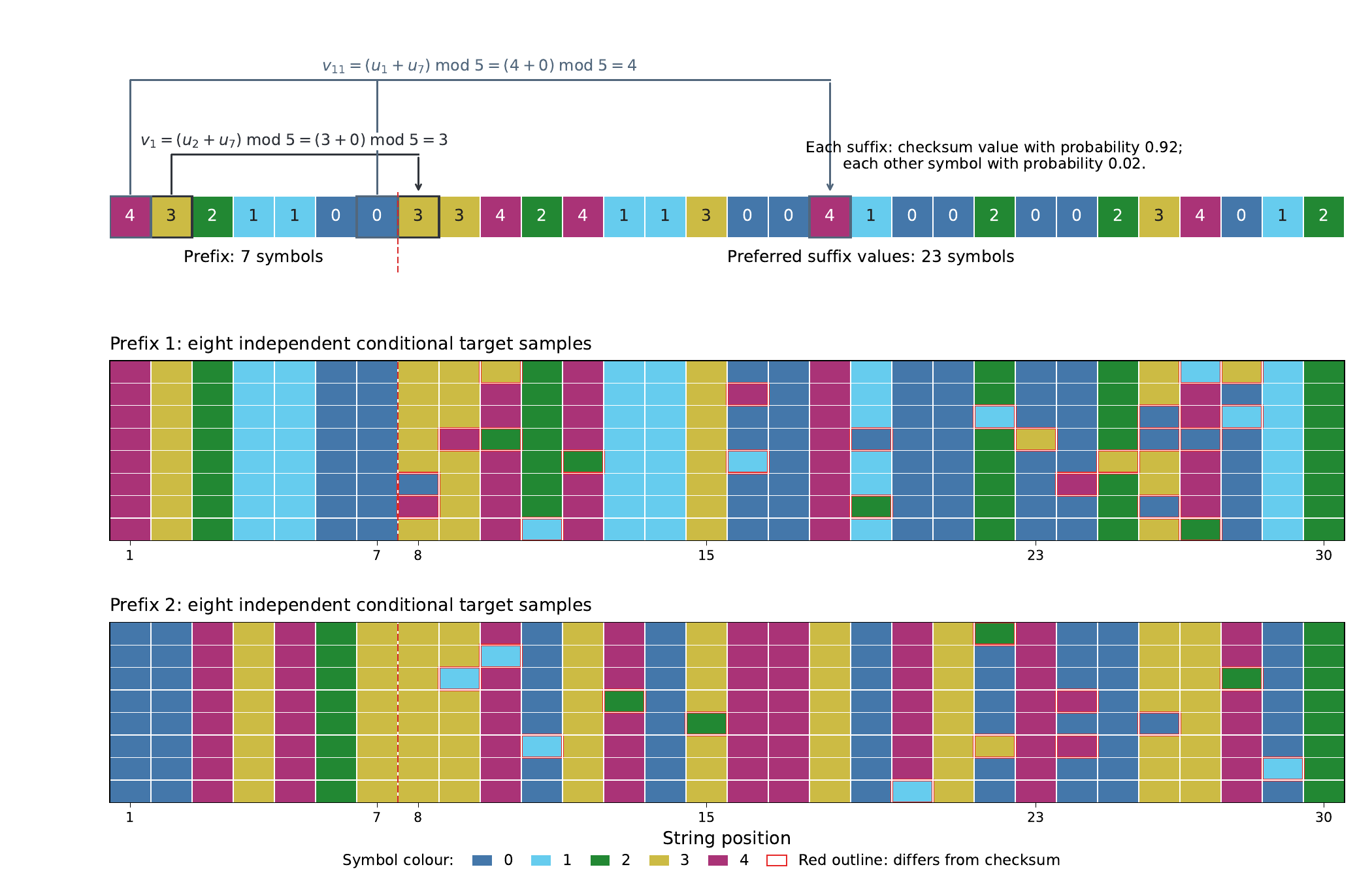}
    \caption{Checksum dependencies with $L=30$, $A=5$, $m=7$, and $\epsilon=0.1$. Top: two actual dependencies from environment seed 0, with arrows connecting prefix inputs to their preferred suffix values. Each suffix symbol equals its checksum with probability $0.92$ and each other value with probability $0.02$. Below: eight independent conditional target samples for each of two uniformly sampled prefixes. Colours encode symbols; red outlines mark deviations from the checksum, and the dashed line separates prefix and suffix.}
    \label{fig:checksum_environment_example}
\end{figure}

\paragraph{Checksum dependencies.}
We use the pairwise half-sum construction with string length $L=30$, alphabet size $A=5$, prefix length $m=7$, and suffix length $r=23$. The prefix symbols are independent and uniform. For each suffix position $j$, two prefix indices $a_j,b_j\in\{1,\ldots,7\}$ are sampled independently with replacement at environment initialisation, so the same prefix position may appear twice in a pair. The pairs are fixed throughout all experiments using environment seed 0, and define $f_j(u)=(u_{a_j}+u_{b_j})\bmod 5$. Conditional on the prefix, suffix symbols are independent: with probability $0.9$ we use $f_j(u)$, and otherwise sample uniformly from all five symbols. Thus, $\epsilon=0.1$, the checksum value has probability $0.92$, and each other symbol has probability $0.02$. We use reward exponent 1, giving the exact normalisation constant $Z=1$. Figure~\ref{fig:checksum_environment_example} illustrates the prefix--suffix dependencies and the variability of completions conditioned on a fixed prefix.

\begin{figure}[htbp]
    \centering
    \includegraphics[width=\linewidth]{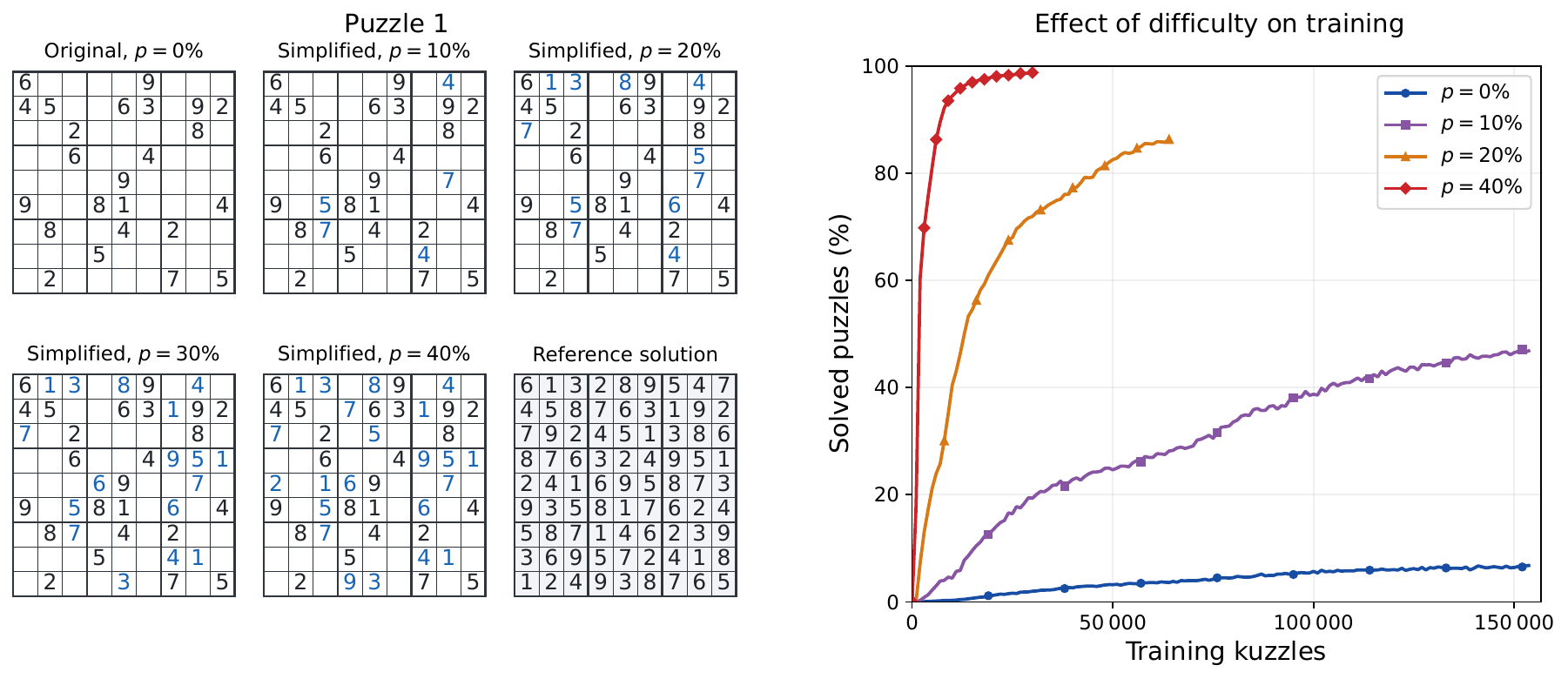}
    \caption{\textbf{Effect of Sudoku difficulty.} \textit{Left}: the same held-out puzzle at $p=0,10,20\%$ (top row), then $p=30,40\%$ and its reference solution (bottom row). Original clues are black and additionally revealed digits are blue. \textit{Right}: raster-order autoregression with the same configuration for $p=0,10,20,40\%$. Solved-puzzle accuracy uses greedy decoding on held-out puzzles at the corresponding difficulty. One kuzzle denotes $1\,000$ training-puzzle presentations.}
    \label{fig:sudoku_difficulty_accuracy}
    \label{fig:sudoku_difficulty_examples}
\end{figure}

\paragraph{Sudoku.}
Our conditional setup uses $\beta=3.75$. We use the puzzles from~\citep{radcliffe33, shah2024causal} and construct easier variants by revealing additional $p$-percent of digits, preserving the original training and test splits. Our order-learning experiments use the $p=10\%$ and $p=30\%$ variants, which reveal approximately one tenth and three tenths of the originally empty cells, respectively. We train on the train split and measure metrics on the test split.
Figure~\ref{fig:sudoku_difficulty_accuracy} illustrates the nested simplifications of one held-out puzzle and compares training at different simplification levels using the same raster-autoregressive configuration. Revealing more digits substantially increases solved-puzzle accuracy at a given training budget.

\clearpage
\subsection{Algorithm hyperparameters}

\paragraph{Shared settings for the string problems.}
We use a Transformer encoder with 3 layers, 8 attention heads, hidden and embedding dimension 64, and feed-forward dimension 64. The dropout rate is 0.1, with no attention dropout. The prediction heads are MLPs with two hidden layers of width 256. In the learned-order model, the symbol, forward-position, and backward-position heads share the forward encoder. The symbol head produces $L\times A$ logits, with a separate categorical symbol distribution for each position; the forward-position head produces $L$ logits. Fixed-order models also predict a separate symbol distribution for every position, selecting the distribution at the prescribed next position.

We optimise the mean squared TB residual with AdamW, using learning rate $10^{-3}$ for the forward encoder and symbol head, $10^{-4}$ as the default backward-head learning rate, weight decay $10^{-3}$, and gradient-norm clipping at 1 within each optimiser parameter group. Adam's moment coefficients are $(0.9,0.999)$ and its numerical stabilizer is $10^{-8}$. The learned scalar $\log Z$ starts at zero and has learning rate $10^{-2}$. These learning rates are constant. Our learned-order recipe uses $\eta_{\mathrm{ord}}=0.1$, giving forward-position and backward-head learning rates $10^{-4}$ and $10^{-5}$, respectively. We initialise the weights and biases of the final layers of both position heads to zero, giving uniform distributions over valid positions; symbol logits retain their default random initialisation.

Each training iteration samples 512 fresh trajectories. Exploration mixes the forward policy with a uniform distribution over valid actions, with its mixture weight decreasing linearly from 0.3 to 0.01 over the first 50\,000 iterations and remaining at 0.01 thereafter. For fixed-order models, exploration changes the symbol but preserves the prescribed position. Where replay is enabled, each iteration additionally performs one separate update on 512 replay trajectories. The buffer holds up to 2\,560\,000 trajectories and is sampled once it contains at least 12\,800. Half of each replay batch is drawn uniformly from the highest-reward 10\% of stored trajectories, and the other half from the remaining trajectories. We do not apply importance weights to replay updates. Thus, an iteration denotes a fresh-data cycle and contains two optimiser updates once replay is active. The branching variant preserves the fresh-trajectory budget, using 128 roots with four continuations each.

\paragraph{Markov-chain strings.}
Our learned-order recipe uses backward-entropy coefficient $\lambda_{H_\text{B}}=3$. We add this positive coefficient times the mean backward-policy entropy over valid trajectory steps to the TB loss, thereby encouraging more deterministic destruction orders. Forward-entropy regularisation is disabled. The fixed-order permutation sweep in Figure~\ref{fig:markov_chain_convergence} uses deterministic backward transitions and no replay buffer; the learned-order ablations use replay unless explicitly stated otherwise. We use ten seeds for the fixed-order sweep and five seeds for each learned-order ablation.

For the fixed-order comparisons, we vary only the generation permutation, using $K\in\{0,5,10,20,40\}$ random swaps of the identity order. Each swap exchanges two distinct positions; positions may be reused in subsequent swaps. For each seed, permutations with larger $K$ extend the same sequence of swaps. Each resulting generation order remains fixed throughout training and evaluation, and backward transitions erase symbols in the reverse order.

\paragraph{Checksum dependencies.}
We use the same architecture, optimiser, exploration schedule, and batch sizes. The learned-order recipe uses $\lambda_{H_\text{B}}=1$, with forward-entropy regularisation disabled. Both the fixed-order sweep and the learned-order experiments use replay. We use ten seeds per fixed-order configuration and five seeds per learned-order configuration. Entropy, initialisation, learning-rate, factorisation, and branching ablations change only their specified components. In the joint position--symbol-logit ablation, the forward head retains its default learning rate and only the backward head receives the $0.1$ multiplier; uniform initialisation then also makes the initial symbol probabilities uniform.

For the fixed-order comparisons, we use $K\in\{0,1,2,3,4,5,6,7\}$. We sample each prefix--suffix swap uniformly among positions not used in previous swaps. For each seed, larger $K$ extends the same sequence of swaps. The resulting order remains fixed throughout training and evaluation, with backward transitions erasing symbols in the reverse order.

\paragraph{Stopping and checkpoints.}
For both tasks, we train for at most 100\,000 iterations and check convergence every 250 iterations. A check estimates each direction of trajectory KL using 2\,048 trajectories. Both estimates must have absolute value below 0.1 for three successive checks. We then validate with 8\,192 trajectories per direction and require each estimate plus twice its standard error to be below 0.1. Since $Z=1$, these criteria correspond to ELBO approaching zero from below and EUBO approaching zero from above. We save resumable checkpoints every 250 iterations and SNR diagnostics every 500 iterations, including a final diagnostic at normal termination. Reported mean convergence iterations assign 100\,000 iterations to runs that do not converge within the budget.

\clearpage
\section{Metrics}
\label{app:metrics}

We evaluate sampling quality using the evidence lower bound (ELBO) and evidence upper bound (EUBO). We follow the evaluation framework of~\citet{blessing2024beyond}, adapted to discrete GFlowNets by~\citet{zykova2026proximal}. These complementary metrics bound the log normalising constant $\log Z$: ELBO evaluates samples generated by the model, while EUBO evaluates coverage of the target distribution.

\paragraph{ELBO.}
For trajectories $\tau=(s_0,\ldots,s_T)$ sampled from the forward policy, the evidence lower bound is
\begin{equation}
    \operatorname{ELBO}
    =\mathbb{E}_{\tau\sim P_F}\left[
    \log\frac{R(s_T)\prod_{t=0}^{T-1}P_B(s_t\mid s_{t+1})}
    {\prod_{t=0}^{T-1}P_F(s_{t+1}\mid s_t)}
    \right]\leq\log Z.
\end{equation}
Higher values indicate better agreement with the target on trajectories visited by the model. However, ELBO can remain high when the model misses target modes, since those modes are rarely sampled by the forward policy. We estimate the expectation by averaging over sampled trajectories.

\paragraph{EUBO.}
We sample terminal states from the target distribution $R(s_T)/Z$ and reconstruct trajectories using the backward policy. The evidence upper bound is
\begin{equation}
    \operatorname{EUBO}
    =\mathbb{E}_{\substack{s_T\sim R(\cdot)/Z\\\tau\sim P_B(\cdot\mid s_T)}}\left[
    \log\frac{R(s_T)\prod_{t=0}^{T-1}P_B(s_t\mid s_{t+1})}
    {\prod_{t=0}^{T-1}P_F(s_{t+1}\mid s_t)}
    \right]\geq\log Z.
\end{equation}
Lower values indicate better coverage of the target distribution. Unlike ELBO, EUBO penalises target trajectories to which the forward policy assigns little probability, making it sensitive to missed modes. We estimate the expectation by averaging over target samples and their sampled backward trajectories.

\paragraph{Importance-weighted EUBO (IW-EUBO).}
EUBO requires terminal samples from the target distribution, which are unavailable for Sudoku. The dataset provides correct solutions, but the reward distribution also assigns probability to incorrect boards. We therefore approximate EUBO using reference solutions and nearby corrupted completions, weighted by their rewards and proposal probabilities.

For each evaluation puzzle, we retain its reference solution and generate two independently corrupted terminal boards. For each corruption, we sample $k$ uniformly from $1,\ldots,\min(20,m)$, where $m$ is the number of originally empty cells. We then choose $k$ distinct such positions uniformly and replace each digit by a uniformly chosen different digit. Original clues remain unchanged. This gives one correct completion and two nearby incorrect completions per puzzle. The corresponding proposal is $q=\tfrac13\delta_{x^*}+\tfrac23q_{\mathrm{corr}}$, where $x^*$ is the reference solution. In particular, $q(x^*)=1/3$, and a board differing from $x^*$ in $k$ eligible cells has probability $q(x)=2/[3\min(20,m)\binom{m}{k}8^k]$.

For a given puzzle, let $\mathcal D$ denote the uniform distribution over its three terminal boards. These boards were constructed using the proposal $q$, rather than sampled from the target distribution. We correct for this difference using importance weights $w(s_T)=R(s_T)/q(s_T)$: dividing by $q(s_T)$ accounts for how frequently the proposal generates each board, while multiplying by $R(s_T)$ favours boards according to the target reward. The resulting estimate is
\begin{equation}
    \operatorname{IW\text{-}EUBO}
    =
    \mathbb{E}_{\substack{s_T\sim\mathcal D\\\tau\sim P_B(\cdot\mid s_T)}}\left[
    w(s_T)\log\frac{R(s_T)\prod_{t=0}^{T-1}P_B(s_t\mid s_{t+1})}
    {\prod_{t=0}^{T-1}P_F(s_{t+1}\mid s_t)} \right]
    \;/\; \mathbb{E}_{s_T\sim\mathcal D}[w(s_T)].
\end{equation}
The denominator normalises the importance weights, so the unknown $Z$ is not needed. We compute these averages over all three boards, sampling a backward trajectory for each, and report the mean of the resulting estimates over all evaluation puzzles. The reference solutions and corruptions are fixed across evaluations. Lower values indicate better coverage of these reward-weighted completions. Unlike EUBO, this metric no longer upper-bounds $\log Z$: self-normalisation introduces finite-sample bias, and the proposal covers only completions within 20 digit changes of the reference solution.

\paragraph{Digit accuracy.}
For Sudoku, we also evaluate the fraction of originally empty cells filled with the correct reference digit. Let $\mathcal V$ be the evaluation set, $B_j$ the set of initially empty cells in puzzle $j$, $x_j^*$ its reference solution, and $\hat{x}_j$ the generated completion. We report
\begin{equation}
    \operatorname{DigitAcc}
    =100\,\frac{\sum_{j\in\mathcal V}\sum_{i\in B_j}
    \mathbf{1}\{\hat{x}_{j,i}=x^*_{j,i}\}}
    {\sum_{j\in\mathcal V}|B_j|}.
\end{equation}
Original clues are excluded, and the counts are pooled across puzzles rather than averaging per-puzzle percentages. Higher values indicate better completion accuracy. For this metric, we decode digits greedily at each chosen position. Ordinary decoding follows the model's generation-order rule; for a learned order, it selects the highest-probability position before selecting the highest-probability digit conditional on that position. Confidence decoding instead selects the empty cell with the largest maximum conditional digit probability. Both variants use the model's own generated digits throughout the trajectory, without reference-digit correction or exploration. Unlike ELBO and IW-EUBO, digit accuracy measures greedy completion quality rather than sampling quality.

\end{document}